\documentclass[11pt]{article}

\usepackage[T1]{fontenc}
\usepackage{lmodern}
\usepackage[utf8]{inputenc}
\usepackage{microtype}
\usepackage{framed}
\usepackage{listings}
\usepackage{vmargin}
\usepackage{setspace}
\usepackage{mathrsfs, mathenv}
\usepackage{amsmath, amsthm, amssymb, amsfonts, amscd}
\usepackage{graphicx}
\usepackage{epstopdf}
\usepackage[svgnames]{xcolor}
\usepackage{hyperref}
\hypersetup{citecolor=blue, colorlinks=true, linkcolor=black}
\usepackage[ruled, vlined]{algorithm2e}
\usepackage[capitalise]{cleveref}
\usepackage{subcaption}
\usepackage{bbm}
\usepackage{cite}
\usepackage{verbatim}
\usepackage{pgfplots}
\usepackage{tikz}
\usetikzlibrary{arrows,decorations.pathmorphing,backgrounds,positioning,fit,matrix}
\usepackage{etoolbox}
\usepackage{color}
\usepackage{lipsum}
\usepackage{ifthen}
\usepackage[title]{appendix}
\usepackage{autonum}
\usepackage{accents}
\usepackage{xpatch}
\usepackage[]{algpseudocode}
\usepackage{multicol}
\usepackage{multirow}
\usepackage[abs]{overpic}
\usepackage{eqnarray}
\usepackage{bm}

\crefname{assumption}{Assumption}{Assumptions}
\crefname{figure}{Figure}{Figures}

\theoremstyle{plain}
\newtheorem{theorem}{Theorem}[section]

\numberwithin{equation}{section}

\theoremstyle{definition}

\theoremstyle{remark}
\newtheorem{remark}[theorem]{Remark}

\ifpdf
  \DeclareGraphicsExtensions{.eps,.pdf,.png,.jpg}
\else
  \DeclareGraphicsExtensions{.eps}
\fi

\usepackage{mathtools}

\usepackage{booktabs}

\usepackage{pgfplots}
\usepackage{tikz}
\usetikzlibrary{patterns,arrows,decorations.pathmorphing,backgrounds,positioning,fit,matrix}
\usepackage[labelfont=bf]{caption}
\usepackage{here}
\usepackage[font=normal]{subcaption}

\usepackage{tablefootnote}

\usepackage{enumitem}
\setlist[itemize]{leftmargin=.5in}
\setlist[enumerate]{leftmargin=.5in,topsep=3pt,itemsep=3pt,label=(\roman*)}

\newcommand{\TheTitle}{
On the performance of multi-fidelity and reduced-dimensional neural emulators for inference of physiologic boundary conditions}
\newcommand{\TheAuthors}{C. H. Choi, A. Zanoni, D. E. Schiavazzi, A. L. Marsden}
\title{\TheTitle}
\author{
Chloe H. Choi \thanks{Department of Mechanical Engineering, Stanford University, Stanford, CA. USA}
\and Andrea Zanoni \thanks{Centro di Ricerca Matematica Ennio De Giorgi, Scuola Normale Superiore, Pisa, Italy.}
\and Daniele E. Schiavazzi \thanks{Department of Applied and Computational Mathematics and Statistics, University of Notre Dame, Notre Dame, IN, USA.}
\and Alison L. Marsden \thanks{Institute for Computational and Mathematical Engineering, Stanford University, Stanford, CA, USA.} \thanks{Bioengineering, Stanford University, Stanford, CA, USA.} \thanks{Pediatric Cardiology, Stanford University, Stanford, CA, USA.} }
\date{}

\newcommand{\Q}{\mathcal Q}
\newcommand{\E}{\mathcal E}
\newcommand{\D}{\mathcal D}

\renewcommand{\S}{\mathcal S}

\newcommand{\LF}{\mathrm{LF}}
\newcommand{\HF}{\mathrm{HF}}

\newcommand{\NF}{\mathrm{NF}}

\usepackage{amsopn}
\DeclareMathOperator{\diag}{diag}

\newcommand{\R}{\mathbb{R}}

\newcommand{\Ex}{\operatorname{\mathbb{E}}}

\newcommand{\dd}{\,\mathrm{d}}
\definecolor{shade}{RGB}{100, 100, 100}
\definecolor{bordeaux}{RGB}{128, 0, 50}

\renewcommand*{\dot}[1]{\accentset{\mbox{\large\bfseries .}}{#1}}

\newcommand{\chloerev}[1]{\textcolor{black}{#1}}

\usepackage[usestackEOL]{stackengine}

\definecolor{leg1}{RGB}{0,114,189}
\definecolor{leg2}{RGB}{217,83,25}
\definecolor{leg3}{RGB}{237,177,32}
\definecolor{leg4}{RGB}{126,47,142}
\definecolor{leg5}{RGB}{119,172,48}

\definecolor{leg21}{RGB}{62,38,169}
\definecolor{leg22}{RGB}{46,135,247}
\definecolor{leg23}{RGB}{55,200,151}
\definecolor{leg24}{RGB}{254,195,56}

\ifpdf
\hypersetup{
	pdftitle={\TheTitle},
	pdfauthor={\TheAuthors}
}
\fi

\begin{document}
	
\maketitle


\begin{abstract}
\noindent Solving inverse problems in cardiovascular modeling is particularly challenging due to the high computational cost of running high-fidelity simulations.
In this work, we focus on Bayesian parameter estimation and explore different methods to reduce the computational cost of sampling from the posterior distribution by leveraging low-fidelity approximations.
A common approach is to construct a surrogate model for the high-fidelity simulation itself. Another is to build a surrogate for the discrepancy between high- and low-fidelity models. This discrepancy, which is often easier to approximate, is modeled with either a fully connected neural network or a nonlinear dimensionality reduction technique that enables surrogate construction in a lower-dimensional space. A third possible approach is to treat the discrepancy between the high-fidelity and surrogate models as random noise and estimate its distribution using normalizing flows. This allows us to incorporate the approximation error into the Bayesian inverse problem by modifying the likelihood function.
We validate five different methods which are variations of the above on analytical test cases by comparing them to posterior distributions derived solely from high-fidelity models, assessing both accuracy and computational cost. 
Finally, we demonstrate our approaches on two cardiovascular examples of increasing complexity: a lumped-parameter Windkessel model and a patient-specific three-dimensional anatomy.

\end{abstract}


\section{Introduction}

With increasing computational power, mathematical models of the cardiovascular system are becoming more accurate and realistic, opening up the possibility of clinical adoption.
Cardiovascular models offer valuable insight for surgical planning, but they are still faced with skepticism in the clinical community due in part to their deterministic nature.
However, recent advancements have opened up new possibilities for uncertainty-aware modeling, including forward propagation, parameter estimation, and end-to-end, data-to-prediction paradigms combining the solution of forward and inverse problems.

High-fidelity cardiovascular models, which are typically full three-dimensional (3D) solutions to the incompressible Navier--Stokes equations, provide detailed information on blood flow dynamics, but are computationally expensive, often requiring hours to days for a single simulation.
Therefore, their use becomes impractical in many-query tasks such as sensitivity analysis, uncertainty quantification, and Bayesian posterior sampling.
To mitigate this issue, reduced-order models (ROMs) are frequently employed as computationally efficient surrogates.
Among these, zero-dimensional (0D) and one-dimensional (1D) models are widely used due to their effective trade-off between computational cost and physiological accuracy.
0D models, also referred to as lumped-parameter models, represent the vasculature as a network of interconnected compartments with spatially uniform pressure and flow, and are governed by systems of ordinary differential equations.
On the other hand, 1D models are obtained via a centerline extraction procedure and retain spatial resolution along the axial direction by averaging the Navier--Stokes equations across vessel cross-sections.
We note that these models can be automatically derived from 3D geometries as described in ~\cite{pfaller22_centerline}.

The problem of parameter estimation is particularly important in cardiovascular applications, where tuning elements such as boundary conditions is essential. 
\chloerev{Deterministic optimization, while valuable, typically seeks only a single point estimate. In contrast, Bayesian inference allows for the quantification of uncertainty in the estimated values by characterizing their full posterior distribution. This is particularly important in clinical settings, where robust predictions and credible intervals are critical for decision-making. Moreover, Bayesian inversion naturally accounts for measurement noise, which is inherent in devices used to collect quantities of interest, such as pressure. Therefore, Bayesian methods are particularly relevant in cardiovascular modeling and can support more informative and realistic clinical decisions.}
Surrogate models have been successfully employed to reduce the computational cost of these inference tasks.
In the context of coronary models, an automated tuning procedure based on surrogate models and derivative-free Nelder--Mead optimization has been proposed in \cite{tran16_automated}.
This framework was then adapted for pulmonary models to tune resistance boundary conditions in \cite{lan22_ppas}.
Moreover, in \cite{richter2024bayesian}, inference via sequential Monte Carlo was demonstrated on 72 models from the Vascular Model Repository \cite{vmr}, a database of cardiovascular computational models, using 0D approximations calibrated against results from a single 3D simulation for each patient.

Replacing high-fidelity 3D models with lower-fidelity approximations can result in a loss of information of the original high-fidelity model. Hence, multilevel and multifidelity strategies have been increasingly adopted in various applications to employ both model types in combination.
\chloerev{Some preexisting multi-fidelity-based approaches include Gaussian progress (GP) regression~\cite{matheron1963principles}-based methods, such as co-Kriging~\cite{revMF_le2014recursive} for thermoelectric generation systems~\cite{revMF_LEE_24}, and hierarchical Kriging~\cite{revMF_han2012hierarchical} for thermal battery design~\cite{revMF_LEE_25} and structural optimization~\cite{revMF_LEE_22}. However, GP-based multi-fidelity methods become computationally expensive and difficult to optimize for higher-dimensional spaces~\cite{perdikaris2017nonlinear, revMF_LIU_18}, and they cannot be readily applied to discontinuous functions~\cite{raissi2016deep}, or inverse problems with strong nonlinearities~\cite{raissi2017inferring}. To mitigate this, artificial neural networks~\cite{revMF_guo2022_ANN, minisci2011robust} and deep neural networks~\cite{meng2020composite} were developed. 
Other dimensionality reduction techniques are explored in~\cite{revMF_BRUNEL_25}. For a comprehensive review of recent multi-fidelity approaches, interested readers can refer to \cite{revMF_fernandez_16, revMF_fernandez_19}.}

In the cardiovascular community, multifidelity Monte Carlo estimators have gained particular interest, where a strong correlation between high- and low-fidelity models has been shown to significantly reduce the variance of estimated quantities of interest through optimal allocation of computational resources \cite{peherstorfer2016optimal}.
An extension of this approach, which couples different levels of discretization accuracy within a multifidelity framework, has been successfully applied to aortic and coronary models to assess the impact of uncertainties in material properties and boundary conditions on model outputs \cite{FLEETER20_mlmf}.
To further enhance the correlation between model fidelities, and thus improve estimator performance, a combination of dimensionality reduction techniques and normalizing flows has been employed in a series of studies \cite{ZGS24b,Z24_improvedMF,ZGSMD24}.
This methodology has also been applied to coronary hemodynamics to predict clinical quantities of interest along with their associated uncertainties \cite{MZK24}.

In this work, we apply \chloerev{the} multifidelity paradigm, which has already been extensively adopted for \chloerev{forward} uncertainty propagation, to tackle Bayesian inverse problems. We compare \chloerev{different} strategies to leverage low-fidelity 0D cardiovascular models to reduce the computational cost of sampling from the posterior distribution, without significantly affecting the accuracy.
\chloerev{We focus on} learning the discrepancy between high- and low-fidelity models, which \chloerev{typically exhibits lower variation, i.e., fewer sharp transitions, making it} more suitable for approximation than the models themselves. To approximate this discrepancy, we employ both fully connected neural networks and reduced-order surrogates based on neural active manifolds (NeurAM~\cite{ZGSMD24}), a recently introduced strategy for nonlinear dimensionality reduction. \chloerev{Specifically, NeurAM learns a low-dimensional nonlinear manifold embedded in the original input space that captures most of the variation in the model. Inputs are then projected onto this manifold, and a surrogate model is learnt only on the manifold.}

Replacing the true model with an approximation inevitably introduces errors in posterior sampling, whose magnitude depends on the approximation error~\cite{2001_discrepancy}. To address this, we propose \chloerev{to} incorporate the modeling error directly into the derivation of the posterior distribution.
Specifically, we treat the discrepancy between the high-fidelity model and its surrogate as a random variable, whose distribution must be inferred.
In this context, prior studies have proposed a dynamic update strategy where the modeling error is assumed to be Gaussian~\cite{Calvetti_2018}, and a similar approach has also been applied in~\cite{zanoni20_enkf} to quantify the difference between multiscale models and their homogenized counterparts in the context of ensemble Kalman filters. Rather than using an online updating strategy, here we adopt an offline approach, where the distribution of the modeling error is first learned using normalizing flows and subsequently used to modify the likelihood function. 
Moreover, we introduce a scaling factor for the surrogate model, which we optimize to minimize the posterior variance.


The main contributions of this work are the following.
\begin{itemize}[leftmargin=*]
    \item In the context of cardiovascular modeling, we evaluate the performance of five different surrogate modeling strategies to reduce the computational cost of posterior sampling in Bayesian inverse problems.
    \item We leverage the discrepancy between high- and low-fidelity models using both fully connected neural networks and lower-dimensional surrogates based on neural active manifolds.
    \item We extend NeurAM to models with multiple outputs.
    \item We employ normalizing flows to estimate the distribution of the modeling error introduced by surrogate replacement and incorporate this distribution into the posterior via a modification of the likelihood function.
    \item We validate the above surrogates on multiple test cases, including maps with closed form expressions and complex cardiovascular models.
\end{itemize}

The remainder of the paper is organized as follows. In \cref{sec:methods}, we introduce the proposed methodologies for reducing the computational cost of Bayesian parameter estimation. Specifically, we begin by summarizing neural active manifolds in \cref{sec:NeurAM}, then present our surrogate modeling and modeling error strategies in \cref{sec:surrogates,sec:model_error}, respectively, and finally revisit Bayesian inverse problems in \cref{Bayesian}. The proposed techniques are first evaluated on analytical test cases in \cref{sec_examples}, and then applied to cardiovascular simulations in \cref{sec:cardio}. We conclude with a discussion of our work and potential future research directions in \cref{sec:conclusion}.

\section{Methodology}\label{sec:methods}

Before discussing the methodology used for the numerical experiments in~\cref{sec_examples}, we briefly introduce the notation. 
Consider a collection of random inputs $\bm{X}$, assumed continuous, with probability distribution $\mu$. 
Also consider a map or \emph{model} $\Q \colon \mathbb{R}^{d} \to \mathbb{R}$, whose evaluations are assumed to be computationally expensive. We emphasize this by saying that this is a \emph{high-fidelity} model and writing $\Q=\Q_{\HF}$.

\subsection{NeurAM for nonlinear dimensionality reduction} \label{sec:NeurAM}

We first introduce a recently proposed data-driven approach for nonlinear dimensionality reduction called \emph{neural active manifolds} or NeurAM~\cite{ZGSMD24}. 
NeurAM maps every location $\bm{x}$ on the support of the random inputs $\bm{X}$ onto a one-dimensional manifold, according to their \emph{level set} $\Q(\bm{x})$.
It consists of an \emph{encoder} $\E \colon \mathbb{R}^{d} \to \mathbb{R}$, projecting the inputs along a one-dimensional latent space, a \emph{decoder} $\D \colon \mathbb{R} \to \mathbb{R}^{d}$ mapping the latent variable back into the original space, and a one-dimensional \emph{surrogate} $\S \colon \mathbb{R} \to \mathbb{R}$.
NeurAM learns a one-dimensional \emph{active} manifold $\D \circ \E$ while simultaneously training a reduced-order surrogate model $\Q_{\S} = \S \circ \E$ by minimizing a loss function of the form
\begin{equation}\label{equ:neuram_loss}
\begin{split}
    \mathcal L(\E,\D,\S) = &\Ex^\mu \left[ (\Q(\bm{X}) - \S(\E(\bm{X})))^2 \right]+\Ex^\mu \left[ (\Q(\bm{X}) - \S(\E(\D(\E(\bm{X})))))^2 \right]+\\
    &\Ex^\mu \left[ (\D(\E(\bm{X})) - \D(\E(\D(\E(\bm{X})))))^2 \right].
\end{split}
\end{equation}
The first and second terms in \eqref{equ:neuram_loss} minimize the loss of the surrogate $\S$ by encoding and re-encoding the model input along the NeurAM, respectively. 
In addition, the third term ensures that a point on the NeurAM is mapped to the point itself after multiple applications of the encoder/decoder pair, or, in other words, the map $\D(\E(\bm{X}))$ is an identity along the NeurAM. 
In practice, the encoder, decoder and surrogate models are all expressed as multilayer perceptrons or MLPs (see, e.g.,~\cite[Chapter 6]{DeepLearning}) and parameterized through weights and biases, i.e., $\E(\cdot,\bm{\alpha})$, $\D(\cdot,\bm{\beta})$, and $\S(\cdot,\bm{\gamma})$, \chloerev{where $\bm{\alpha}, \bm{\beta}$, and $\bm{\gamma}$ represent the network parameters.} Optimal values of these parameters are determined \chloerev{by minimizing the loss function in equation \eqref{equ:neuram_loss}} through gradient descent using Adam~\cite{adam_opt_2017}. 
For further information on NeurAM and its performance on a wide variety of test cases, the interested reader is referred to~\cite{ZGS24}.

To accommodate models with multiple outputs, the dimension of the latent space is selected to be equal to the number of outputs, and the dimensions are learned jointly within a single shared model. 
We demonstrate this strategy for a Windkessel model and an aorto-iliac bifurcation example, where the diastolic, systolic, and mean pressures must all be considered.

\subsection{Surrogate likelihood functions}\label{sec:surrogates}

In this section, we consider five different surrogate models, which we refer to as Methods B--F, that are used to produce an approximate Gaussian likelihood, as further discussed in~\cref{Bayesian}.
These include a $d$-dimensional dense neural network surrogate directly estimated from the high-fidelity model. We refer to a surrogate of this kind as $\widetilde{\Q} \simeq \Q_\HF$.
In addition, we consider cases where a computationally inexpensive \emph{low-fidelity} model $\Q_{\LF}$ is available which provides a possibly biased approximation of $\Q_{\HF}$. For such cases, we consider surrogates of the difference or \emph{discrepancy} between the high- and low-fidelity models, which we refer to as $\widetilde{\Delta} \simeq \Q_\HF - \Q_\LF$. 

\chloerev{We remark that, while a low-fidelity model can be biased, it may still exhibit high correlation with respect to the high-fidelity model, as discussed in~\cite{kupper1984effec,amouzgar2017radial,peherstorfer2018survey}.}
\chloerev{Acceptable Pearson correlation coefficients are typically $0.8$ or higher. This, combined with a computationally inexpensive low-fidelity model, is enough to improve the robustness of statistical moment estimates in uncertainty propagation outer-loop tasks, see, e.g., \cite[Section 3.4]{peherstorfer2016optimal}.}
\chloerev{In this context, NeurAM was designed to construct a low-fidelity model re-parameterization leading to increased Pearson correlations. However, in the context of Bayesian inference, unlike in uncertainty propagation, a highly correlated surrogate model does not necessarily lead to improved posterior distributions. Bayesian inference is also sensitive to the bias of the surrogate model, which can significantly affect the accuracy of the posterior estimate.}

We finally consider one-dimensional surrogates computed using the NeurAM framework introduced in the previous section. Consistent with the notation in~\cite{ZGS24}, we use $\widetilde{\Q}_{\S}$ and $\widetilde{\Delta}_{\S}$ for one-dimensional surrogates which refer directly to the high-fidelity quantity of interest or the low- to high-fidelity discrepancy, respectively.

Based on these surrogates, we compare methods that differ based on how the model output is approximated. Specifically, considering a statistical model of the form
\begin{equation}
    y = \Q_{\text{HF}}(\bm{x}) + \eta, \quad \text{where} \quad \eta \sim \mathcal{N}(0, \sigma_{\text{noise}}^2),
\end{equation}
\chloerev{where $\mathcal N(m, \Sigma)$ denotes a Gaussian random variable with mean $m$ and variance $\Sigma$}, the following methods are used to construct a likelihood function $\rho(y|\bm{x})$.

\begin{itemize}[leftmargin=*]
\item {\bf Method A} - In this case, we directly use the high-fidelity model without any approximation
\begin{equation}
    \rho_A(y|\bm{x}) = \frac{1}{\sqrt{2\pi\sigma_{\text{noise}}^2}} \text{exp} \left\{ -\frac{1}{2 \sigma_{\text{noise}}^2} \left[y-\Q_\HF(\bm{x})\right]^2 \right\},
\end{equation}
and therefore refer to this case as the \emph{ground truth}.
\item {\bf Method B} - We use a Gaussian likelihood, where the high-fidelity model is replaced by a surrogate 
\begin{equation}
    \rho_B(y|\bm{x}) = \frac{1}{\sqrt{2\pi\sigma_{\text{noise}}^2}}  \text{exp} \left\{ -\frac{1}{2 \sigma_{\text{noise}}^2} \left[y-\widetilde{\Q}(\bm{x})\right]^2 \right\}.
\end{equation}
\item {\bf Method C} - We formulate the likelihood using the low-fidelity model and train a surrogate model to learn the discrepancy between the high- and low-fidelity models. This leads to the approximation
\begin{equation}
    \rho_C(y|\bm{x}) = \frac{1}{\sqrt{2\pi\sigma_{\text{noise}}^2}} \text{exp} \left\{ -\frac{1}{2 \sigma_{\text{noise}}^2} \left[y-(\Q_{\text{LF}}(\bm{x}) + \widetilde{\Delta}(\bm{x}))\right]^2 \right\}.
\end{equation}
\item {\bf Method D} - We use NeurAM to compute a surrogate for the high-fidelity model, leading to
\begin{equation}
    \rho_D(y|\bm{x}) = \frac{1}{\sqrt{2\pi\sigma_{\text{noise}}^2}} \text{exp} \left\{ -\frac{1}{2 \sigma_{\text{noise}}^2} \left(y-\widetilde{\Q}_{\S}(\bm{x})\right)^2 \right\}.
\end{equation}
\item {\bf Method E} - We evaluate the low-fidelity model and train NeurAM to learn the discrepancy between the high- and low-fidelity models, leading to an approximate likelihood of the form
\begin{equation}
    \rho_E(y|\bm{x}) = \frac{1}{\sqrt{2\pi\sigma_{\text{noise}}^2}} \text{exp} \left\{ -\frac{1}{2 \sigma_{\text{noise}}^2} \left[y-(\Q_{\text{LF}}(\bm{x}) + \widetilde{\Delta}_{\S}(\bm{x})) \right]^2 \right\}.
\end{equation}
\end{itemize}

\begin{remark}
    All the methods discussed above, as well as in the following presentation, are described for the case of a scalar output and single observation. Nevertheless, they can be easily extended to handle multiple quantities of interest in $\R^m$ and $K$ \chloerev{independent} observations $\{ \bm{y^{(k)}} \}_{k=1}^K$. In such cases, we assume a multivariate Gaussian noise distribution $\boldsymbol\eta \sim \mathcal N(0, \bm{\Sigma}_{\text{noise}})$, along with its associated probability density function. For example, for Method A, we have the following expression: 
    \begin{equation}
    \rho_A(\bm{y}|\bm{x}) = \prod_{k=1}^K\frac1{(2\pi)^{m/2} \sqrt{\det(\bm{\Sigma}_\text{noise})}} \exp \left\{ -\frac{1}{2} \left[\bm{y}^{(k)}-\Q_\HF(\bm{x})\right]^\top \bm{\Sigma}_{\text{noise}}^{-1} \left[\bm{y}^{(k)}-\Q_\HF(\bm{x})\right] \right\},
\end{equation}
where $\det(\bm{\Sigma}_\text{noise})$ denotes the determinant of the noise covariance matrix, and 
\begin{equation}
    \bm{y}^{(k)} = \begin{bmatrix} y_1^{(k)} & \cdots & y_m^{(k)} \end{bmatrix}^\top,
\end{equation}
is the $k$-th $m$-dimensional observation.
\end{remark}

\subsection{Modeling error}\label{sec:model_error}

In this section, we propose an alternative strategy that accounts for the error introduced by replacing the original high-fidelity model with a surrogate. Let $Q^\dagger$ denote a surrogate model for $\Q_\HF$, which can be any approximation including those introduced in Methods B, C, D, E above. Then, we write for all $\alpha \in \R$
\begin{equation} \label{eq:methodF_general}
    y = \Q_\HF(\bm{x}) + \eta = \alpha Q^\dagger(\bm{x}) + \underbrace{\Q_{\HF}(\bm{x}) - \alpha Q^\dagger(\bm{x}) + \eta}_{\widetilde{\eta}},
\end{equation}
which can be seen as an inverse problem for the model $\alpha Q^\dagger$ with noise given by $\widetilde \eta$ incorporating both the original noise and the modeling error. The advantage of this approach is that the high-fidelity model $Q_\HF$ appears only in the noise term $\widetilde \eta$, eliminating the cost of evaluating it during posterior sampling. However, this introduces a new challenge because the noise distribution is now unknown and no longer Gaussian. To address this, we propose to use normalizing flows~\cite{kobyzev2020normalizing,papamakarios2021normalizing}, in particular consisting of a collection of real-valued  non-volume preserving transformations (RealNVP~\cite{DSB17,SLC23}), to estimate the distribution of an \emph{inflated noise} $\widetilde{\eta}\sim\widetilde\rho_{\text{NF}}(\bm{x})$.

RealNVP-based normalizing flows consist of $L$ bijective transformations
\begin{equation}
    \mathcal G(\bm{z}; \bm{\vartheta}) = (g_L(\cdot; \bm{\theta}_L) \circ \cdots \circ g_2(\cdot; \bm{\theta}_2) \circ g_1(\cdot; \bm{\theta}_1))(\bm{z}),
\end{equation}
parameterized by $\bm{\vartheta} = (\bm{\theta}_1, \dots, \bm{\theta}_L)$, such that $\widetilde\rho_\mathrm{NF} = \mathcal G(\cdot;\bm{\vartheta})_\#\,\mathcal N(0,\bm{I})$, where the subscript $\#$ denotes the push-forward measure. Under this formula, the density $p$ of the standard Gaussian is transformed by the change of variable formula into
\begin{equation}
    \widetilde{\rho}_\mathrm{NF}(\bm{\Delta}) = p(\bm{Z})\,|\mathcal G'(\bm{Z};\bm{\vartheta})|^{-1},
\end{equation}
where $\bm{z} = \mathcal G^{-1}(\bm{\delta};\bm{\vartheta})$, or, equivalently,
\begin{equation}
    \log \widetilde{\rho}_\mathrm{NF}(\bm{\delta}) = \log p(\bm{z}_{0}) - \sum_{\ell=1}^L \log |g_\ell'(\bm{z}_{\ell-1}; \bm{\theta}_\ell)|,
\end{equation}
where $\bm{z} = \bm{z}_{0} = \mathcal G^{-1}(\bm{\delta};\bm{\vartheta})$ and $\bm{z}_{\ell} = g_\ell(\bm{z}_{\ell-1};\bm{\theta}_\ell)$. The parameters $\bm{\vartheta}$ are then obtained by maximizing the log-likelihood of the available samples $\{ \bm{\delta}^{(j)} \}_{j=1}^J$
\begin{equation}
    \sum_{j=1}^J \log \widetilde\rho_\mathrm{NF}(\bm{\delta}^{(j)}) = \sum_{j=1}^J \log p(\bm{z}_{0}^{(j)}) - \sum_{j=1}^J \sum_{\ell=1}^L \log |g_\ell'(\bm{z}^{(j)}_{\ell-1}; \bm{\theta}_\ell)|,
\end{equation}
where $\bm{\delta}^{(j)} = \Q_\HF(\bm{x}^{(j)}) - \alpha Q^\dagger(\bm{x}^{(j)})$, $\bm{z}^{(j)}_{0} = \mathcal G^{-1}(\bm{\delta}^{(j)};\bm{\vartheta})$, and $\bm{z}^{(j)}_{\ell} = g_\ell(\bm{z}^{(j)}_{\ell-1};\bm{\theta}_\ell)$. 
In general, since the variance of $\widetilde{\eta}$ is greater than the variance of $\eta$ because the modeling error $\Q_\HF - \alpha Q^\dagger$ is independent of $\eta$, the variance of the resulting posterior distribution will also be overestimated. To mitigate this phenomenon, we select an optimal coefficient $\alpha$ resulting from the minimization of the variance of $\widetilde\eta$, that is
\begin{equation} \label{eq:alpha_opt}
    \alpha_{\text{opt}} = \frac{\mathbb{C}^{\mu}\left[\Q_{\HF}(\bm{X}), Q^\dagger(\bm{X})  \right]}{\mathbb{V}^{\mu}\left[Q^\dagger(\bm{X}) \right]},
\end{equation}
whose derivation is provided in Appendix \ref{app:alpha}, \chloerev{and where $\mathbb{C}^\mu(\cdot)$ and $\mathbb{V}^\mu(\cdot)$ denote the covariance and variance operators with respect to the probability distribution $\mu$ of the random input $\bm{X}$, respectively.}

\begin{figure}[!htb]
    \centering\includegraphics[width=0.85\linewidth]{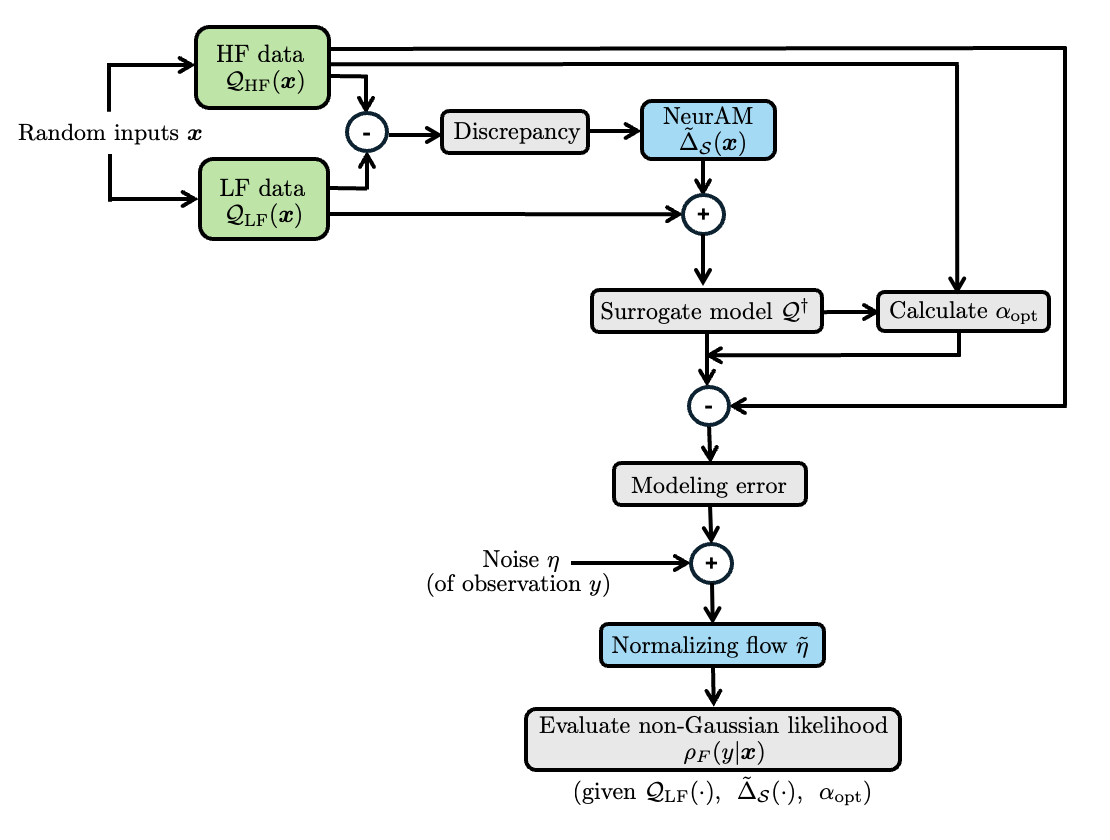}
    \caption{\chloerev{Flow chart of Method F, whose key elements include main steps (gray), training data (green), and neural networks for nonlinear dimensionality reduction and density estimation (blue).}}
    \label{fig:methodF_workflow}
\end{figure}

We adopt this approach, where $Q^\dagger$ is replaced by the surrogate model $(\Q_\LF + \widetilde \Delta_\S)$ used in Method E, for use in the final method that we test in the numerical experiments.
\begin{itemize}[leftmargin=*]
\item {\bf Method F} - Evaluate the non-Gaussian likelihood as
\[
\rho_F(y|\bm{x}) = \widetilde\rho_{\text{NF}} \left\{ y - \alpha_{\text{opt}} \left[ \Q_{\LF}(\bm{x}) + \widetilde{\Delta}_{\S}(\bm{x}) \right] \right\}.
\]
\end{itemize}
We note that the main difference between Method F and the previous methods lies in the likelihood formulation, which is no longer Gaussian due to the inclusion of the surrogate model error. We expect this to reduce the bias of the posterior distribution at the price of increasing its variance. Therefore, this approach is particularly suitable for scenarios where the primary goal is to obtain the maximum a posteriori estimate, and where an overestimation of the posterior variance is acceptable. \chloerev{A flow chart for Method F is shown in Figure \ref{fig:methodF_workflow}}.

\chloerev{The specific steps of Method F, and its comparison to Methods B--E, are highlighted in \cref{alg:method_f}. In terms of contribution, Methods C through F are all new, each introducing a different form of novelty, while Method B simply uses a fully connected neural network to learn a surrogate of the high-fidelity model. In particular, Method D replaces the fully connected neural network with NeurAM, a novel nonlinear dimensionality reduction technique, and Methods C and E aim to learn the model discrepancy. Finally, Method F leverages normalizing flows to account for the distribution of the modeling error in the likelihood function. Then, the main goal of this paper is to compare these methods and highlight situations where performance trade-offs may occur.}

\begin{algorithm}
    \caption{\chloerev{Construction of posterior distribution from surrogate model (Methods B--F) and density estimator (Method F).}} \label{alg:method_f}
    \begin{algorithmic}[1]
        \State \chloerev{\textbf{Input:} $N$ random inputs, their corresponding high-fidelity model evaluations, $\mathcal{Q}_{\text{HF}}(\boldsymbol{x})$, low-fidelity model evaluations, $\mathcal{Q}_{\text{LF}}(\boldsymbol{x})$, as well as variance $\sigma_{\text{noise}}^2$ of the measurement noise $\eta$.
        \State \textbf{Initialization:} Initialize surrogate models for the different methods.}

        \chloerev{Methods B, C: Fully-connected neural network.}

        \chloerev{Methods D, E, F: NeurAM.}
        
        \State \chloerev{\textbf{Training:} Train surrogate model on input-output pairs split into 75\% training, 25\% testing.}
        
        \chloerev{Methods B and D: train on $\boldsymbol{x}$ and $\mathcal{Q}_{\text{HF}}(\boldsymbol{x})$.}
        
        \chloerev{Methods C, E, and F: train on $\boldsymbol{x}$ and $\mathcal{Q}_{\text{HF}}(\boldsymbol{x}) - \mathcal{Q}_{\text{LF}}(\boldsymbol{x})$.}

        \State \If{\chloerev{Method F}}

        \chloerev{Construct surrogate model $Q^\dagger(\bm{x}) = \mathcal{Q}_{\LF}(\boldsymbol{x})+\tilde{\Delta}_{\mathcal{S}}(\bm{x})$.}
        
        \chloerev{Compute the optimal coefficient $\alpha_{\text{opt}}$.}

        \chloerev{Train RealNVP $\rho_F(y|\boldsymbol{x})$ on $\mathcal{Q}_{\text{HF}}(\boldsymbol{x}) - \alpha_{\text{opt}}\mathcal{Q}^{\dagger}(\boldsymbol{x}) + \eta$.}
        
        \Return \chloerev{posterior distribution combining density estimator $\rho_F(y|\boldsymbol{x})$ and surrogate model.}

        \State \Else

        \Return \chloerev{posterior distribution combining multivariate Gaussian likelihood and surrogate model.}
        
    \end{algorithmic}
    \end{algorithm}

\subsubsection{Rationale for method F}

\begin{figure}[!htb]
    \centering
    \includegraphics[width=\linewidth]{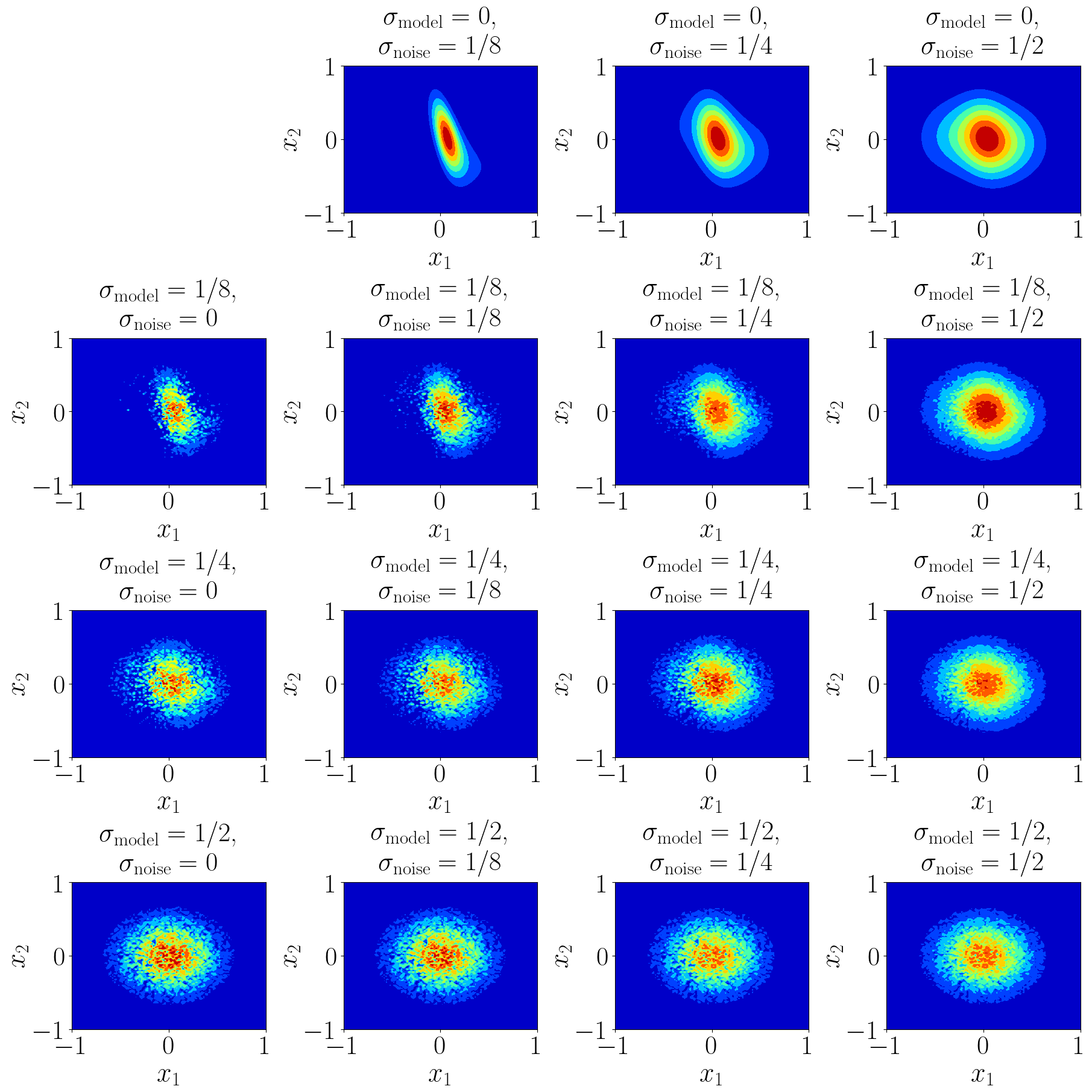}
    \caption{Posterior distributions obtained using Method F varying $\sigma_\text{model}$ and $\sigma_\text{noise}$. Here the low-fidelity model is a spatially uncorrelated perturbed version of the high-fidelity model.}
    \label{fig:uncor_noise_grid}
\end{figure}

From the discussion in the previous section, Method F is expected to have superior performance for cases where the low- to high-fidelity model discrepancy can be effectively modeled as noise, particularly when the distribution of the latter can be easily expressed through normalizing flows.
Therefore, we examine the case where this discrepancy is exactly Gaussian, or, in other words, the low-fidelity model is a spatially uncorrelated noise-corrupted version of the high-fidelity model.
To further analyze this case, we examine the effect of different levels of model error and observation noise on the performance of method F.
Consider a model pair with inputs in two dimensions $\bm{x}=(x_1,x_2) \in \R^2$ defined as
\begin{align}
    \Q_{\text{HF}}(\bm{x}) &= e^{0.7x_1 + 0.3x_2} + 0.15\sin(2\pi x_1),\\
    \Q_{\text{LF}}(\bm{x}) &= Q_{\text{HF}}(\bm{x}) + \varepsilon_{\text{model}}(\bm{x}), \,\, \text{where} \,\, \varepsilon_{\text{model}}(\bm{x}) \sim \mathcal{N}(0, \sigma_{\text{model}}^2),
\end{align}
where we have high confidence that the low-fidelity model is discontinuous at every point.
We simplify the setup in equation~\eqref{eq:methodF_general} replacing $\alpha_\text{opt}$ by $\alpha=1$ and without training a surrogate model $\widetilde\Delta_\S$, i.e., $\widetilde\Delta_\S = 0$ or equivalently $Q^\dagger = \Q_\LF$. Then, we obtain
\begin{equation}
    y = \Q_\LF(\bm{x}) + \widetilde\eta, \quad \text{where} \quad \widetilde\eta = \eta - \varepsilon_{\text{model}}(\bm{x}).
\end{equation}
Therefore, since $\varepsilon_{\text{model}}$ is independent of $\eta$, the exact distribution of the inflated noise $\widetilde\eta$ is
\begin{equation}
\widetilde\eta \sim \mathcal N(0, \sigma_{\text{noise}}^2 + \sigma_{\text{model}}^2),
\end{equation}
which we assume to be unknown and we infer using normalizing flows. Let $\widetilde\rho_\mathrm{NF}$ be the estimated probability density function of $\widetilde\eta$ and let $\pi_\text{prior}$ be a bidimensional Gaussian prior distribution for the Bayesian inverse problem (see next section)
\begin{equation}
    \pi_{\text{prior}} \sim \mathcal N(\bm{0}, \bm{\Sigma}), \quad \text{where} \quad \bm{\Sigma} = \begin{bmatrix}
    1/9 & 0 \\ 0 & 1/9
    \end{bmatrix}.
\end{equation}
In \cref{fig:uncor_noise_grid} we visualize on the grid $[-1,1]^2$ the posterior distribution $\widetilde\rho_\mathrm{NF}(y - \Q_\LF(\bm{x})) \pi(\bm{x})$ obtained varying the standard deviation of both the model error, \chloerev{$\sigma_{\text{model}} = \{0, 1/8, 1/4, 1/2\}$}, and the noise, \chloerev{$\sigma_{\text{noise}} = \{0, 1/8, 1/4, 1/2\}$}. We use as observation $y=1.1297$ corresponding to the point $x=(0.5211, 0.2038)$. The first row, where $\sigma_\text{model}=0$, represents the exact posterior distribution given by the original high-fidelity model and corresponds to Method A. As $\sigma_\text{model}$ increases, the discrepancy between the low-fidelity and high-fidelity models becomes more significant. Consequently, directly replacing $\Q_\HF$ with $\Q_\LF$ to reduce the computational cost would result in an inaccurate posterior distribution. The goal of our approach is to transfer the error introduced by the model approximation into the variance of the posterior distribution. This leads to a less precise, but more reliable, quantification of the unknown parameter. Moreover, in the plots in \cref{fig:uncor_noise_grid}, we observe that the overall shape of the posterior distributions, and therefore the behavior of the two sources of uncertainties, remains similar regardless of whether $\sigma_\text{model}$ or $\sigma_\text{noise}$ is increased. However, in the presence of model error, the posterior distributions lose their smoothness, since $\Q_\LF$ is discontinuous at every point.

\begin{figure}[!htb]
    \centering
    \includegraphics[width=\linewidth]{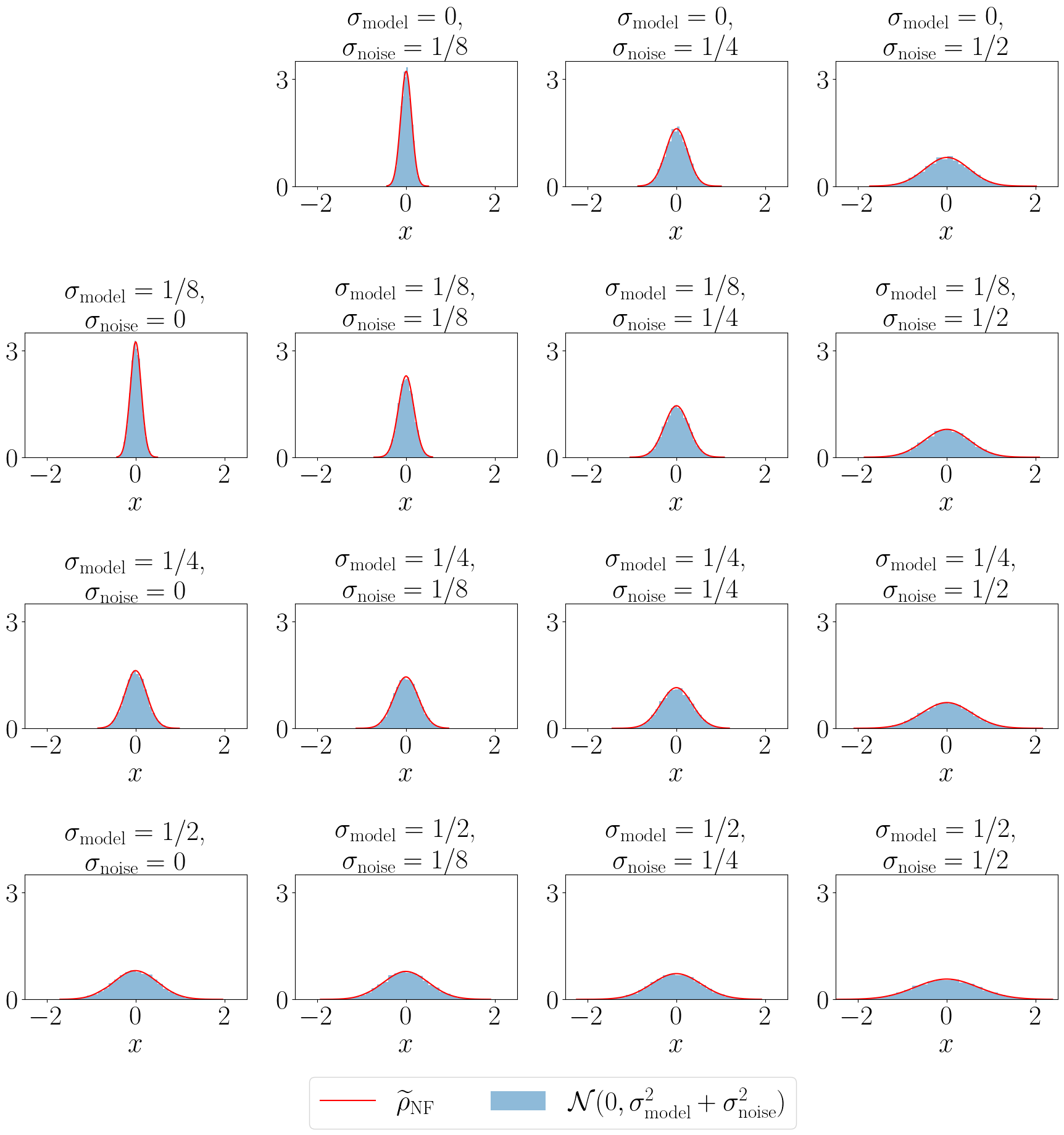}
    \caption{Normalizing flow estimation of the probability density of the noise $\widetilde\eta$ in Method F. Here the low-fidelity model is a spatially uncorrelated perturbed version of the high-fidelity model.}
    \label{fig:uncor_noise_diff}
\end{figure}

In \cref{fig:uncor_noise_diff}, we also plot the histogram of 10,000 realizations of the noise $\widetilde\eta$, along with its approximate density obtained using RealNVP. We construct a RealNVP with 4 layers, 8 neurons, 4 coupling blocks, learning rate of 1e-3, scheduler step of 0.9999, and 5000 epochs. Each data set size is 10,000 split into 25\% for testing and 75\% for training. We observe that normalizing flows accurately capture the underlying distribution.

\subsection{Bayesian inversion}\label{Bayesian}

The solution of inverse problems is a fundamental problem in the analysis of computational models under uncertainty, where the distribution of the model inputs is \emph{inferred} from noisy observations. 
Assume we observe the output of a computationally expensive model corrupted by a zero-mean Gaussian noise with variance $\sigma_{\text{noise}}^2$, or, in other words, we adopt a statistical model of the form
\begin{equation}\label{equ:stat-model}
    y = \Q_{\text{HF}}(\bm{x}) + \eta, \quad \text{where} \quad \eta \sim \mathcal{N}(0, \sigma_{\text{noise}}^2).
\end{equation}
The selected noise model in~\eqref{equ:stat-model} induces a Gaussian likelihood,
\[
\rho(y|\bm{x}) = \mathcal{N}(\Q_{\text{HF}}(\bm{x}),\sigma_{\text{noise}}^2),
\]
where, by an abuse of notation, we do not distinguish between the measure and its associated probability density function.
We also account for prior knowledge on the inputs $\bm{X}$ in the form of a \emph{prior distribution} $\pi(\bm{x})$ on the space of the model inputs. In all numerical test cases in \cref{sec_examples,sec:cardio} we consider a uniform prior 
\[
\pi(\bm{x}) = \mathcal{U}([a_1,b_1]\times[a_2,b_2]\times\cdots\times[a_d,b_d]),
\]
where $[a_{i},b_{i}]$ represent an admissible prior range in the $i$-th dimension for all $i = 1, \dots, d$. 
The celebrated Bayes theorem suggests that \emph{conjunction} of information from the likelihood and prior can be accomplished through simple multiplication, or 
\begin{equation}
    \rho(\bm{x}|y) = \frac{\rho(y|\bm{x})\,\rho(\bm{x})}{\rho(y)} \propto \rho(y|\bm{x})\,\rho(\bm{x}) = \rho(y|\bm{x})\,\pi(\bm{x}),
\end{equation}
where, depending on the argument, $\rho$ denotes the marginal or conditional probability density functions. We evaluate the above posterior using two different strategies. 
First, $\rho(\bm{x}|y)$ can be visualized after being systematically evaluated over a Cartesian (tensor product) grid. We use this procedure in the analytical test cases in \cref{sec:analytical_example,sec:Michalewicz}.
This provides a straightforward approach to compare the posterior produced by all the likelihood approximations discussed above. The distribution produced by each approach is then compared with the reference solution with uniform prior given by
\begin{equation}
    \rho(\bm{x}|y) = \frac{1}{C}  \text{exp} \left\{ -\frac{1}{2 \sigma_\text{noise}^2}\left[y-\Q_{\text{HF}}(\bm{x})\right]^2 \right\} \mathbbm{1}_{[a_1,b_1]\times[a_2,b_2]\times\cdots\times[a_d,b_d]}(\bm{x}),
\end{equation}
where $\mathbbm{1}_{A}$ denotes the indicator function of set $A$, $C$ is the appropriate normalization constant, $\sigma_\text{noise}$ is the noise standard deviation, $y$ the available observation, and $\Q_{\HF}(\bm{x})$ the high-fidelity model evaluated at the location of interest. 

Second, we might want to generate samples from $\rho(\bm{x}|y)$. This is not, in general, a trivial task as the resulting posterior might be multimodal or characterized by a complex support in high dimensions. We adopt a DREAM sampler~\cite{vrugt_2009} to perform this task, as discussed in Section~\ref{sec:circuits}.
Independent of which approach is selected, each evaluation of $\Q_{\HF}$ is computationally expensive, leading to intractable inverse problems.
For this reason, surrogate models are often used to effectively reduce the computational cost of inference. 

\section{Numerical examples}\label{sec_examples}

In this section, we first evaluate the performance of the strategies introduced in the previous section using \chloerev{three} analytical test cases: a simple test function, the Michalewicz function, \chloerev{and the borehole function. The first two examples} allow us to compute the ground truth and easily visualize the posterior distributions through two-dimensional plots. More complex numerical experiments involving cardiovascular simulations are presented in \cref{sec:cardio}.

\chloerev{We compute the Pearson correlation coefficients for each of the following examples and organize them in \cref{tab:pearson_corr_coeffs}.
%
%
For the two-dimensional analytical cases, we compute the Pearson correlation coefficients on a $100\times100$ grid, while for the higher-dimensional problems, we uniformly sample 10,000 points and evaluate the models on those points. For the last cardiovascular example, the correlation is computed using only the available 25 samples. We find that the surrogate models improve the correlation remarkably in contrast to the baseline low-fidelity model, particularly for the two-dimensional analytical examples, where the initial correlation is poor.}

\begin{table}[!htb]
\caption{\chloerev{Pearson correlation coefficients for the test cases discussed in the paper. Specifically, a $100 \times 100$ uniform grid is used for two-dimensional test cases, whereas $10,000$ points selected uniformly at random are used for higher dimensional problems. For the aortic bifurcation test case, the correlation is computed using 25 testing samples.}}
\resizebox{\textwidth}{!}{
\begin{tabular}{c c c}
\toprule
{\bf Problem, Input dim} & {\bf Method} & {\bf Pearson correlation coefficient} \\
\midrule
Analytical, 2D & Low-fidelity & 0.41417 \\
& B & $0.99958 \pm 0.00000$ \\
& C & $0.97929 \pm 0.00207$  \\
& D & $0.99936 \pm 0.00002$ \\
& E & $0.98120 \pm 0.00142$ \\
\midrule
Michalewicz, 2D & Low-fidelity & 0.73372 \\
& B & $0.92207 \pm 0.05385$ \\
& C & $0.88756 \pm 0.09929$ \\
& D & $0.87705 \pm 0.06483$ \\
& E & $0.76117 \pm 0.05975$ \\
\midrule 
Borehole, 8D & Low-fidelity & 0.99999 \\
& B & $0.99925 \pm 0.00032$ \\
& C & $0.99996 \pm 0.00001$ \\
& D & $0.99876 \pm 0.00058$ \\
& E & $0.99995 \pm 0.00003$ \\
\midrule
Circuit, 3D & Low-fidelity & [0.98654,  0.97246, 0.99730] \\
& B & $[0.99975 \pm 0.00005, 0.99979 \pm 0.00005, 0.99990 \pm 0.00002]$ \\
& C & $[0.99989 \pm 0.00004, 0.99994 \pm 0.00002, 0.99998 \pm 0.00002]$ \\
& D & $[0.99950 \pm 0.00001, 0.99955 \pm 0.00001, 0.99994 \pm 0.00001]$ \\
& E & $[0.99994 \pm 0.00001, 0.99995 \pm 0.00002, 0.99998 \pm 0.00001]$ \\ 
\midrule
Aortic bifurcation, 2D & Low-fidelity & [0.99999 0.99999 0.99999] \\
& B & [0.99389 0.99983 0.99987] \\
& C & [0.98313 0.99984 0.99999] \\
& D & [0.99391 0.99975 0.99989] \\
& E & [0.98345 0.99984 0.99999] \\
\bottomrule
\end{tabular}
}
\label{tab:pearson_corr_coeffs}
\end{table}

\subsection{Two-dimensional analytical function} \label{sec:analytical_example}

\begin{figure}[!htb]
    \centering
    \includegraphics[width=0.6\linewidth]{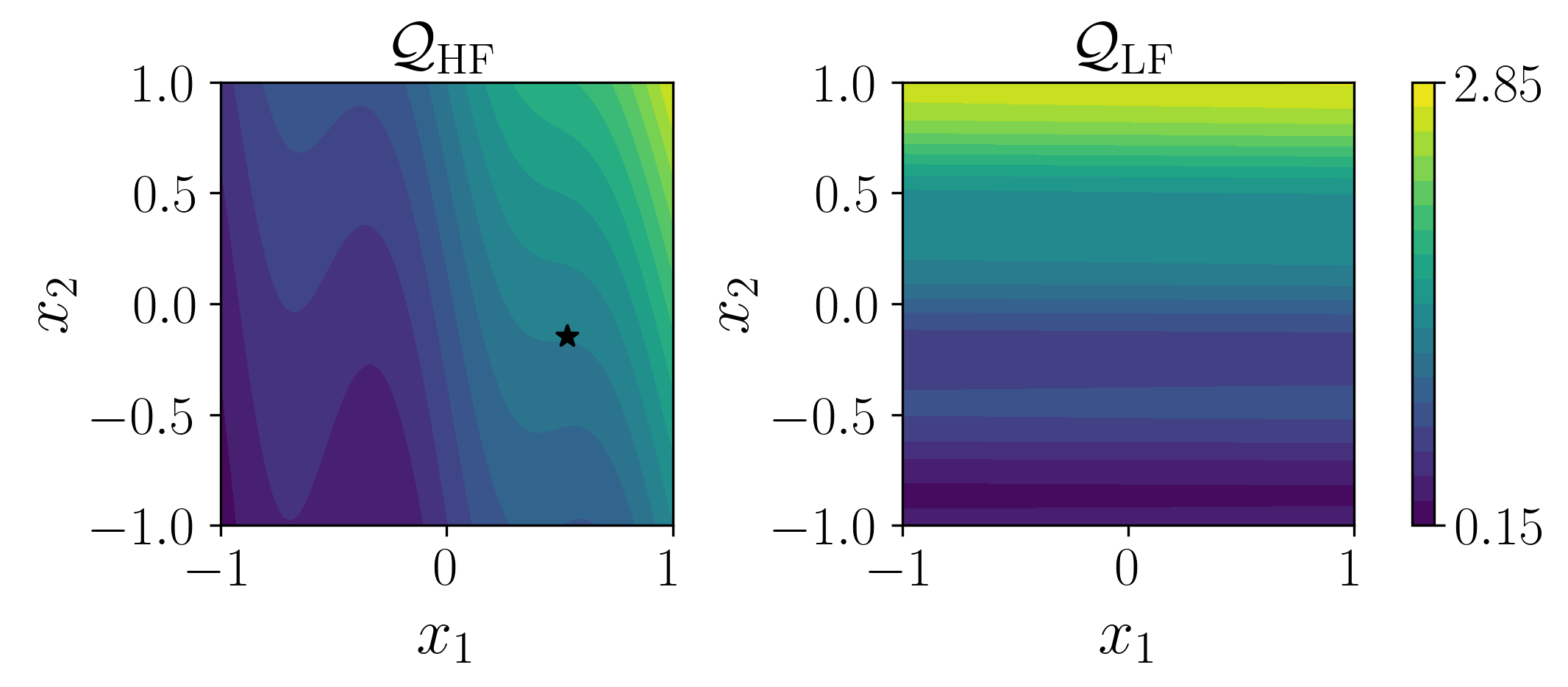}
    \caption{High- and low-fidelity models for the two-dimensional analytical example. The \emph{true} combination of parameters is marked with a black star on the left plot. \chloerev{Contour limits shown on right.}}
    \label{fig:analytical}
\end{figure}

Consider a high- and low-fidelity model pair expressed as
\begin{align}
\Q_\HF(\bm{x}) &= e^{0.7x_1 + 0.3x_2} + 0.15\sin(2\pi x_1), \\
\Q_\LF(\bm{x}) &= e^{0.01x_1 + 0.99x_2} + 0.15\sin(3\pi x_2),
\end{align}
where $\bm{x}=(x_1,x_2)\in [-1, 1]^2\subset\R^{2}$.
This pair, which is poorly correlated, was previously used in~\cite{GEG18} to test the application of active subspaces to multifidelity UQ, and further employed in~\cite{ZGSMD24} and~\cite{ZGS24} in the context of nonlinear dimensionality reduction. 
The desired HF and LF models are shown in Figure~\ref{fig:analytical}. 
We use $\sigma_{\text{noise}}=1/10$.

We show the results for all five Methods B to F outlined in \cref{sec:methods} using a fixed computational budget $N = 100$, and compare the solution against Method A.
The performance of each method is assessed by computing the Hellinger distance $H$ between the resulting posterior distribution and the high-fidelity baseline (Method A). 
Recall that for two continuous probability density functions $\rho_1$ and $\rho_2$ with support in $\R^d$ the Hellinger distance is given by
\begin{equation}
    H(\rho_1, \rho_2) = \sqrt{\frac12 \int_{\R^d} \left( \sqrt{\rho_1(\bm{x})} - \sqrt{\rho_2(\bm{x})} \right)^2 \dd \bm{x}},
\end{equation}
and it holds $H(\rho_1, \rho_2) \in [0,1]$.

\begin{figure}
    \centering
    \begin{subfigure}[t]{0.75\textwidth}
        \centering
        \includegraphics[width=0.8\linewidth]{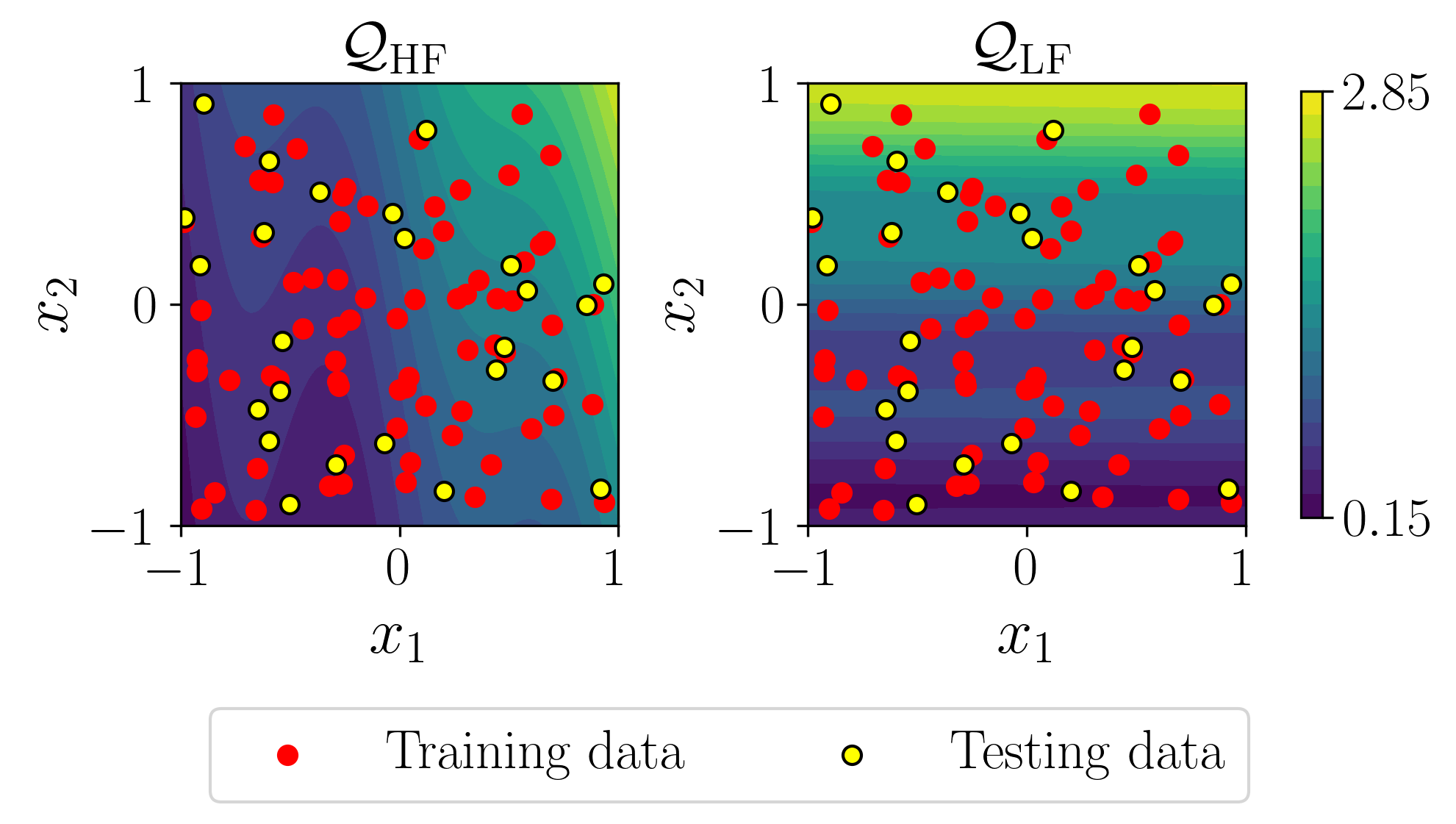}
        \caption{Training and testing data overlaid on the high- and low-fidelity models. \chloerev{Contour limits shown on right.}}
    \end{subfigure} 
    
    \vspace{1em}
    
    \begin{subfigure}[t]{0.85\textwidth}
        \centering
        \includegraphics[width=\textwidth]{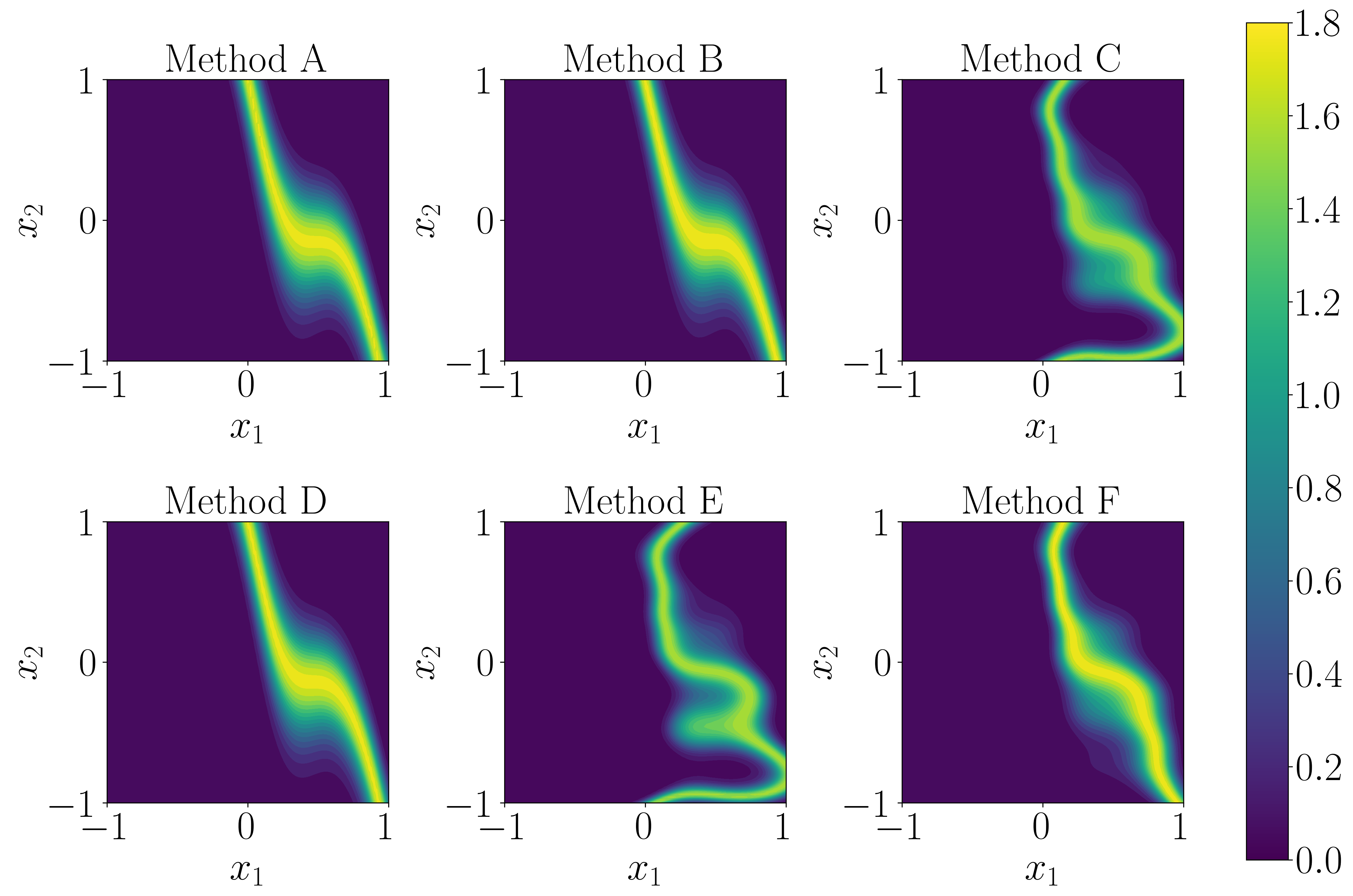}
        \caption{Posterior distributions computed using Methods A to F.}
    \end{subfigure}
    
    \caption{Posterior distributions for the two-dimensional analytical example. The surrogate models for Methods A to F are built using $N=100$ data points and evaluated on $100 \times 100$ grid points.}
    \label{fig:analytical_n_100}
\end{figure}

The results for $N=100$ are shown in~\cref{fig:analytical_n_100}. 
Panel (a) shows the data points split into 75/25 for training/testing. \chloerev{We use uniform sampling to generate the dataset for the benchmark function discussed in this section. Additionally, results for Latin hypercube sampling~\cite{santner2019space} are reported in~\cref{app:analytical_LHS}.}
Note that the observed output $y=1.3547$ corresponds to the input location $[x_1, x_2] = [0.5328, -0.1466]$. 
We remark that, since we selected a noninformative uniform prior distribution, the problem is nonidentifiable (there is a one-dimensional manifold that contains equally likely, maximum a posteriori parameter combinations), but all six methods are successful in capturing the level set containing the true inputs.
We further observe in \cref{tab:analytical_N_100} that the Hellinger distances for Methods B and D are the smallest, i.e., for this simple analytical example with relatively sparse data ($N=100$), the true posterior from Method A is best captured by neural networks (either a fully connected neural network or NeurAM) trained directly on the high-fidelity model.
We also note that the computational times reported in \cref{tab:analytical_N_100} are larger than those for Method A, because in this example the HF model is an analytical function, which is inexpensive to evaluate. 
Therefore, from a computational time point of view, there is no advantage in replacing it with a low-fidelity or surrogate model.
Note that training times are significantly affected by hyperparameter optimization, which we perform using the Optuna library~\cite{Optuna}, with additional details discussed in the Appendix. 

Moreover, accounting for the discrepancy between the high-fidelity and low-fidelity models (Methods C and E) is less effective in these test cases. This is primarily because the low-fidelity model differs significantly from the high-fidelity one, so modeling the discrepancy offers no advantage. 
On the other hand, despite a larger Hellinger distance, Method F appears to slightly improve upon Method E when considering the overall shape of the posterior distribution, as shown in panel (b) of \cref{fig:analytical_n_100}.
We finally note that by increasing the number of data points to $N=500$, we obtain accurate approximations of the posterior distributions for all methods, as shown in \cref{fig:N500_app} in \cref{app:analytical_N_500}.
 
\begin{table}[!ht]
\centering
\caption{Accuracy, variance, and computational cost of the posterior distributions obtained using $N=100$ samples, for the two-dimensional analytical example. Results are averaged over 10 independent experiments. ``Posterior'' refers to the time spent to evaluate the posterior distribution on a $100 \times 100$ grid, one point at a time.}
\begin{tabular}{c c c c c}
\toprule
\multirow{2}{*}{\bf Method} & \multirow{2}{*}{\bf Hellinger distance} & \multirow{2}{*}{tr($\mathbb{C}^{\mu}(\cdot)$)} & \multicolumn{2}{c}{\bf Cost}\\
  & & & {\bf Optuna} & {\bf Posterior}\\
\midrule
 {\bf A} &  0.0 & $0.290$ & - & $0.55$ s\\
 {\bf B} &  $\boldsymbol{0.0259 \pm 0.009}$ & $0.292 \pm 0.004$ & $829 \pm 151$ s & $1.3 \pm 0.3$ s\\
 {\bf C} &  $0.260 \pm 0.073$ & $0.320 \pm 0.051$ & $\boldsymbol{726 \pm 133}$ s &  $ 1.4 \pm 0.3$ s\\
 {\bf D} &  $0.032 \pm 0.015$ & $\boldsymbol{0.291 \pm 0.003}$ & $2118 \pm 579$ s & $\boldsymbol{1.2 \pm 0.4}$ s\\
 {\bf E} &  $0.265 \pm 0.062$ & $0.326 \pm 0.032$ & $2112 \pm 521$ s & $1.4 \pm 0.3$ s\\
 {\bf F} &  $0.281 \pm 0.064$ & $0.347 \pm 0.044$ & $4258 \pm 996$ s & $14.3 \pm 9.2$ s\\
 \bottomrule
\end{tabular}
\label{tab:analytical_N_100}
\end{table}

\subsection{Michalewicz function} \label{sec:Michalewicz}

As a second example, we test the Michalewicz function, a benchmark problem commonly used to evaluate the performance of optimization algorithms, including Bayesian optimization~\cite{jcp}. 
The Michalewicz function is particularly challenging due to its steep valleys and ridges, as well as the presence of many local minima. 
In particular, the regions with steep gradients occupy only a small fraction of the support, making the global minimum difficult to locate when only a limited number of samples are available. 
Moreover, its difficulty is easily adjustable, as the number of local minima increases with the dimensionality, and the steepness of the valleys and ridges depends on a tunable parameter. 
Given a number $d$ of dimensions, the Michalewicz function is defined as
\begin{equation} \label{eq:mich}
    f_m(\bm{x}) = - \sum_{i=1}^d \sin(x_i)\left[\sin(i\,x_i^2/\pi)\right]^{2m},\,\,\text{for}\,\,\bm{x} \in [0,\pi]^d,
\end{equation}
where $m$ is the parameter controlling the steepness of the valleys and ridges. We choose $d=2$ so that both the functions and the posterior distributions can be easily visualized, and assume $\sigma_{\text{noise}}=0.05$.

\begin{figure}[!htb]
    \centering
    \includegraphics[width=0.6\linewidth]{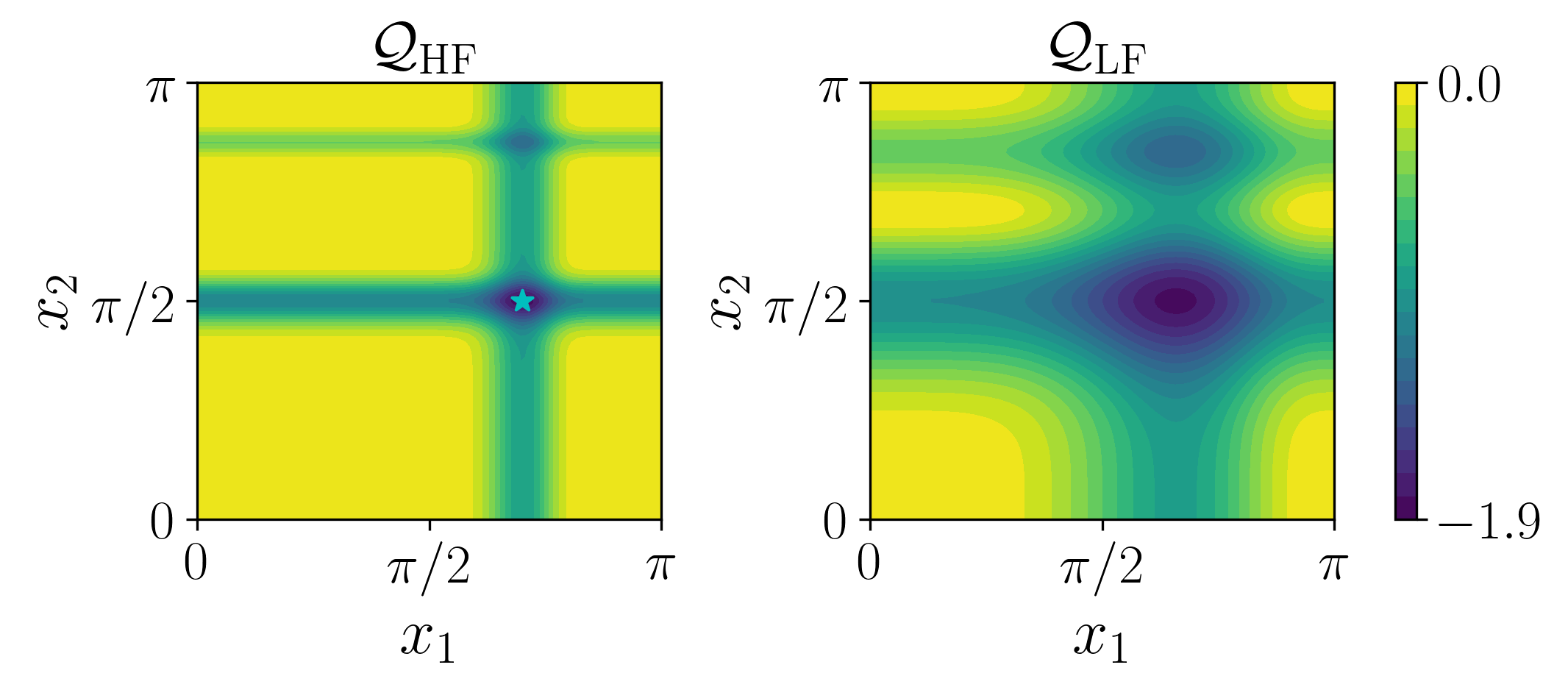}
    \caption{High- and low-fidelity models for the Michalewicz function test case. The \emph{true} parameter combination is marked with a cyan star in the left plot. \chloerev{Contour limits shown on right.}}
    \label{fig:michalewicz}
\end{figure}

We also define the high-fidelity and low-fidelity models by selecting different values of the parameter $m$. 
Specifically, we choose $m=10$ for the high-fidelity model and $m=1$ for the low-fidelity model, i.e.
\begin{equation}
\Q_\HF(\bm{x}) = f_{10}(\bm{x}) \qquad \text{and} \qquad \Q_\LF(\bm{x}) = f_1(\bm{x}),
\end{equation}
with two-dimensional plots shown in~\cref{fig:michalewicz}.

\begin{figure}[!htb]
    \centering
    \begin{subfigure}[t]{0.75\textwidth}
        \centering
        \includegraphics[width=0.8\linewidth]{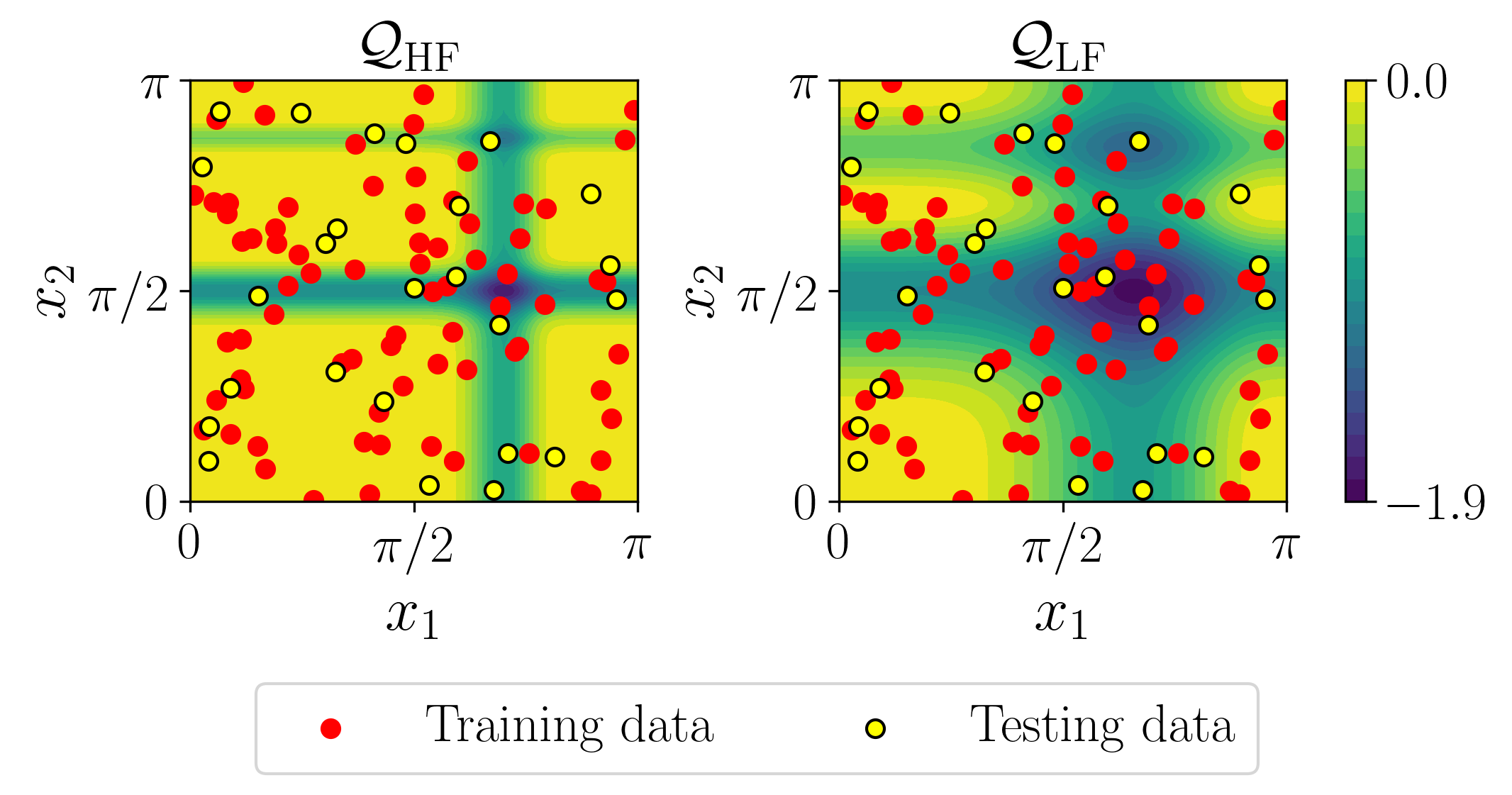}
        \caption{Training and testing data overlaid on the high- and low-fidelity models. \chloerev{Contour limits shown on right.}}
    \end{subfigure} 
    
    \vspace{1em}
    
    \begin{subfigure}[t]{0.85\textwidth}
        \centering
        \includegraphics[width=\textwidth]{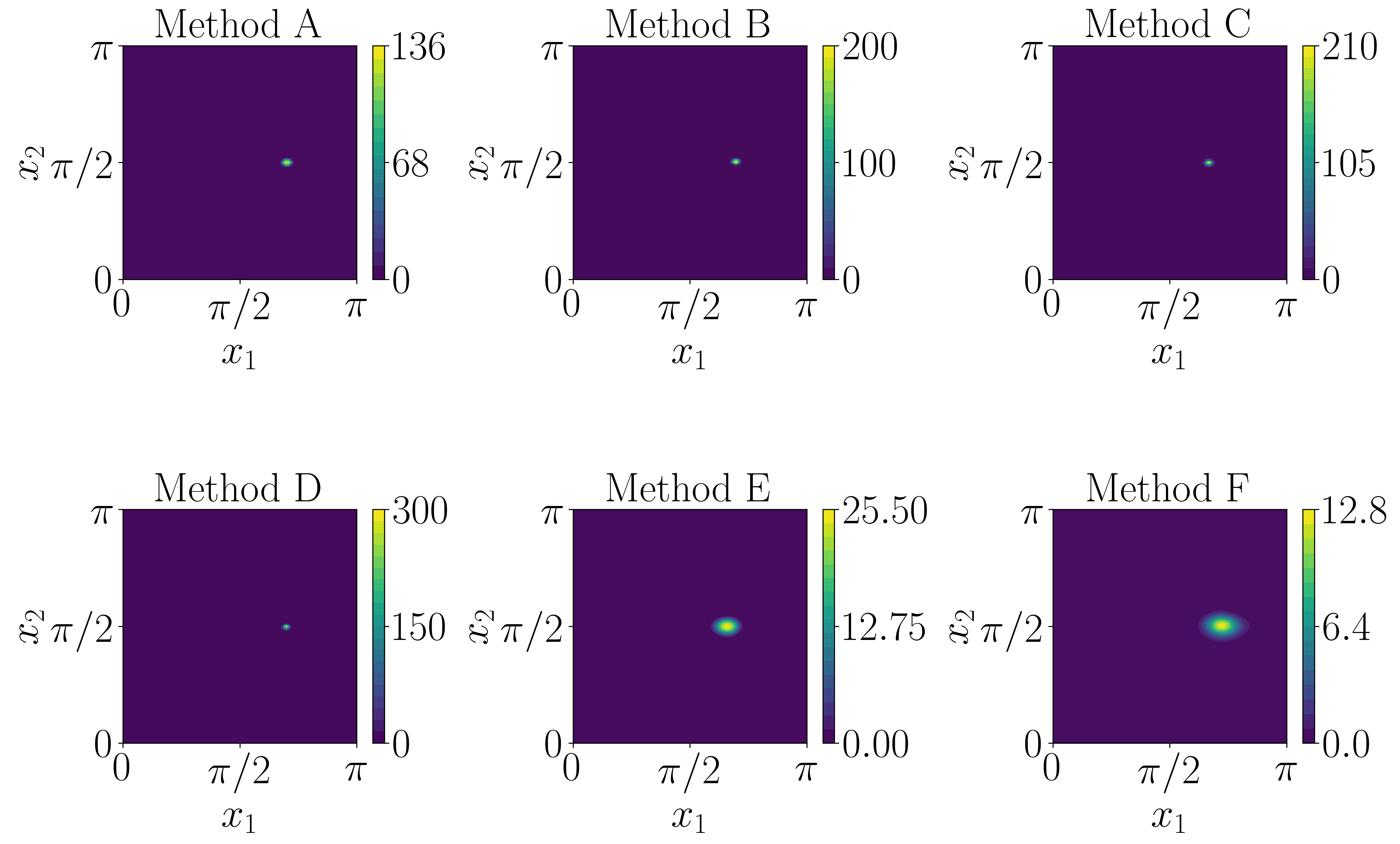}
        \caption{Posterior distributions obtained using the six different methods.}
    \end{subfigure}
    
    \caption{Posterior distributions for the Michalewicz function test case, where the surrogate models are built using $N=100$ data and evaluated on $100 \times 100$ grid points.}
    \label{fig:michalewicz_n_100}
\end{figure}

\chloerev{We uniformly sample $N=100$ data  split into 75/25  training/testing as shown in Panel (a) of \cref{fig:michalewicz_n_100}. Additionally, we demonstrate the effect of Latin hypercube sampling in \cref{app:analytical_LHS}.} We use the observation $y=1.8133$ corresponding to a true input parameter of $\bm{x}=(2.20, 1.57).$ We display the results in~\cref{fig:michalewicz_n_100,tab:grad_N_100}, where we observe that qualitatively the six methods provide a reliable approximation of the true posterior distribution. The Hellinger distances reported in \cref{tab:grad_N_100} indicate that Method B achieves the highest accuracy. For this test case, Methods E and F are more robust, since the standard deviation of the Hellinger distance is significantly smaller, and Method F shows a slight improvement over Method E. 
%

\begin{table}[!ht]
\centering
\caption{Accuracy, variance, and computational cost of the posterior distributions obtained using $N=100$ samples, for the Michalewicz function test case. Results are averaged over 10 independent experiments.}
\resizebox{\textwidth}{!}{
\begin{tabular}{c c c c c}
\toprule
\multirow{2}{*}{\bf Method} & \multirow{2}{*}{\bf Hellinger distance} & \multirow{2}{*}{tr($\mathbb{C}^{\mu}(\cdot)$)} & \multicolumn{2}{c}{\bf Cost}\\
& & & {\bf Optuna} & {\bf Posterior}\\
\midrule
{\bf A} &  0.0 & 0.001786 & - & 0.55 s\\
{\bf B} &  $\bm{0.429 \pm 0.217}$ & $0.001567 \pm 0.000681$ & $493 \pm 72$ s & $\boldsymbol{0.8 \pm 0.2}$ s\\
{\bf C} &  $0.511 \pm 0.284$ & $\bm{0.001279 \pm 0.000428}$ & $\boldsymbol{486 \pm 71}$ s & $ 1.1 \pm 0.2 $ s\\
{\bf D} &  $0.514 \pm 0.275$ & $0.00292 \pm 0.00221$ & $1534 \pm 236$ s & $0.9 \pm 0.1$ s\\
{\bf E} &  $0.842 \pm 0.071$ & $0.00684 \pm 0.00440$ & $1582 \pm 293$ s & $1.3 \pm 0.2$ s\\
{\bf F} & $0.770 \pm 0.066$ & $0.0304 \pm 0.0236$ & $3658 \pm 543$ s & $14 \pm 10$  s\\
\bottomrule
\end{tabular}}
\label{tab:grad_N_100}
\end{table}

\subsection{\chloerev{Borehole function}}

\chloerev{In this section, we test a hydrological model proposed in \cite{harper1983sensitivity}, with eight inputs and one single output representing the flow of water through a borehole. The Borehole function has been studied previously for uncertainty quantification \cite{geraci2025enabling}, active learning \cite{morris1993bayesian, joseph2008blind, gramacy2012gaussian, xiong2013sequential}, and Bayesian experimental design \cite{blanchard2021output}. The original model, which we use as the high-fidelity function, is given by:}
\begin{equation}
    \chloerev{\mathcal{Q}_{\text{HF}}(\boldsymbol{x}) = \frac{2 \pi T_u (H_u - H_l)}{\ln(r/r_w)\Big( 1 + \frac{2LT_u}{\ln(r/r_w)r_w^2K_w}+\frac{T_u}{T_l}\Big)},}
\end{equation}
\chloerev{where}
\begin{align}
    \chloerev{r_w} &\in \chloerev{[0.05, 0.15]}, &
    \chloerev{r} &\in \chloerev{[100, 50000]}, &
    \chloerev{T_u} &\in \chloerev{[63070, 115600]}, &
    \chloerev{H_u} &\in \chloerev{[990, 1110]}, \\
    \chloerev{T_l} &\in \chloerev{[63.1, 116]}, &
    \chloerev{H_l} &\in \chloerev{[700, 820]}, &
    \chloerev{L} &\in \chloerev{[1120, 1680]}, &
    \chloerev{K_w} &\in \chloerev{[9855, 12045]}.
\end{align}
\chloerev{The lower-fidelity model is represented in
\cite{xiong2013sequential} as:}
\begin{equation}
    \chloerev{\mathcal{Q}_{\text{LF}}(\boldsymbol{x}) = \frac{5\, T_u (H_u - H_l)}{\ln(r/r_w)\Big( 1.5 + \frac{2LT_u}{\ln(r/r_w)r_w^2K_w}+\frac{T_u}{T_l}\Big)}.}
\end{equation}
\chloerev{We generate 100 synthetic observations following the univariate distribution $\mathcal{N}(m, \sigma_{\text{noise}}^{2})$, where $m$ is the output value $y=70.9223$ evaluated at the midpoint of the previously mentioned domain bounds and $\sigma_{\text{noise}} = 0.7$.}

\chloerev{Additionally, we use a truncated multivariate normal as the prior distribution:}
\begin{equation}
    \chloerev{\pi(\boldsymbol{x}) \sim \mathcal{N}(\boldsymbol{m},\mathbf{\Sigma}, \boldsymbol{a}, \boldsymbol{b}),}
\end{equation}
\chloerev{where $\boldsymbol{m}$ represents the \emph{true} underlying inputs located at the center of the prior range specified above, with upper and lower bounds in $\boldsymbol{a}$ and $\boldsymbol{b}$. We also assume $\mathbf{\Sigma}=\text{diag}(\sigma^2\,\boldsymbol{I}_{8})$, where $\boldsymbol{I}_{8}$ is the eight-dimensional identity, and consider three different levels for the prior, i.e., $\sigma=\{10^{0.01}, 10^{0.1}, 10^{0.5}\}$. Denoting $\sigma_{\text{log}}=\log_{10}\sigma$, we can also express such levels as $\sigma_{\text{log}} = \{0.01,0.1,0.5\}$.
We compare the resulting posterior distribution from each prior uncertainty with that arising from the uniform distribution to observe trends in identifiability due to the use of informative priors. We sample the resulting posterior distribution with \texttt{PyDREAM} for 50,000 iterations using 5 chains until we reach a Gelman-Rubin statistic of 1.01. 
We note here that the use of a truncated normal Gaussian prior in cardiovascular applications has been previously investigated, such as for inference in a mathematical model of the pulmonary circulation~\cite{puaun2018mcmc}.}

\begin{figure}[!htb]
    \centering
    \includegraphics[width=\linewidth]{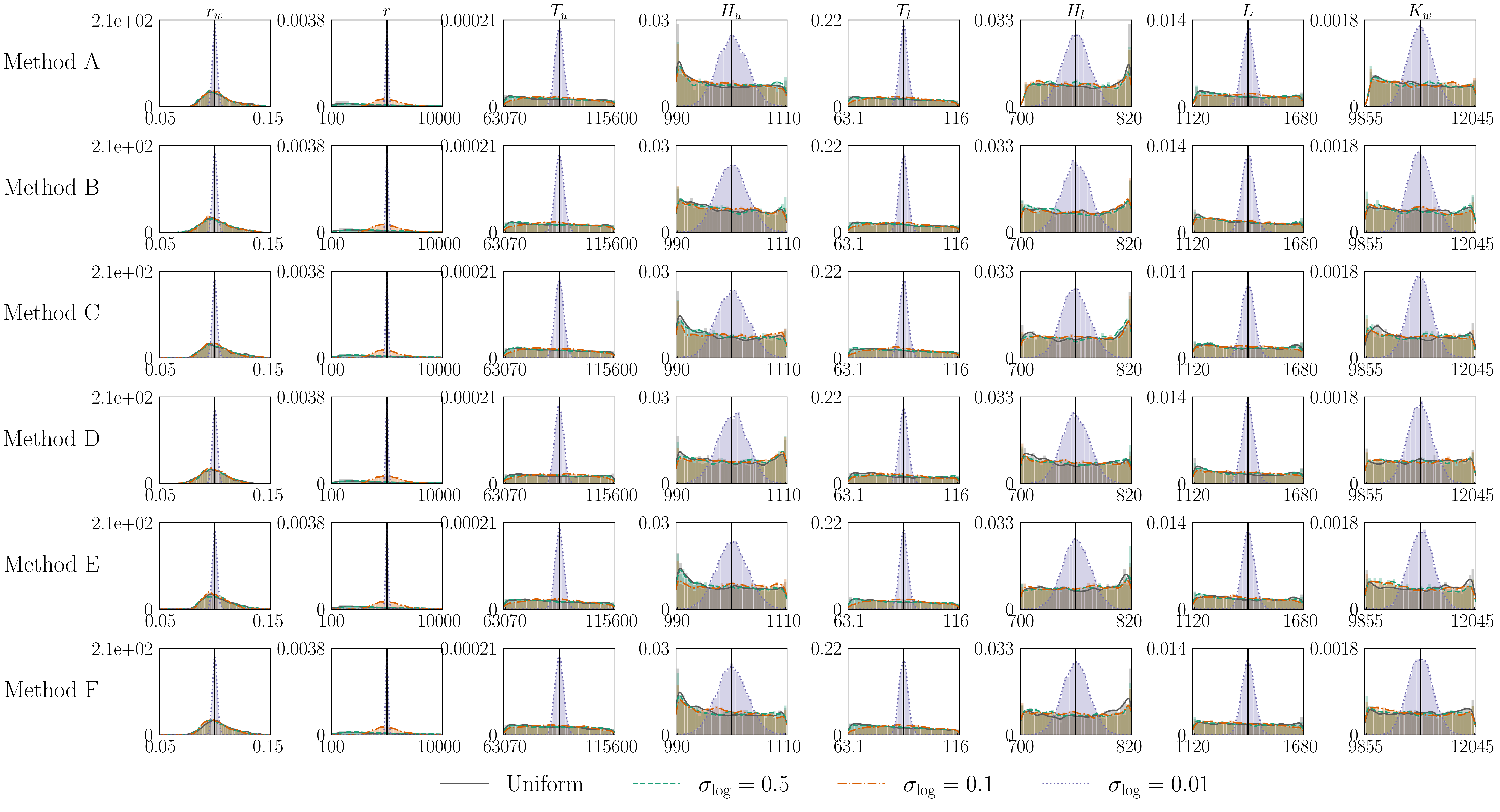}
    \caption{\chloerev{Histograms and KDE curves of the resulting marginal distributions corresponding to each parameter of the borehole function}}
    \label{fig:borehole_histograms}
\end{figure}

\chloerev{We display the resulting posterior distributions in \cref{fig:borehole_histograms}. We observe that the posterior distributions are nearly identical to each other. For better visibility, we include kernel density estimation (KDE) on the sample histograms obtained from running \texttt{PyDREAM}.}

\chloerev{For quantitative comparison, we compile the results in \cref{tab:borehole_N_100_uniform,tab:borehole_N_100_gaussian_0p01,tab:borehole_N_100_gaussian_0p1,tab:borehole_N_100_gaussian_0p5}. We observe similar KL divergences for the various methods. Furthermore, the discrepancy-based methods (Methods C and E) improve KL divergence more than the high-fidelity data-based methods (Methods B and C). We also see that for $\sigma_{\text{log}}=0.5$ and $\sigma_{\text{log}}=0.1$, Method F performs the best, while for $\sigma_{\text{log}}=0.01$ (informative Gaussian prior) and uniform distribution, the discrepancy methods (either C or E) perform the best. Regarding variance, since the input parameters have different ranges and magnitudes, we report the trace of the rescaled covariance, using the midpoint of each parameter’s admissible range as a reference value. Specifically, letting $\boldsymbol{M} = \diag(\boldsymbol{m})$, we rescale the covariance matrix of the posterior distributions by pre- and post-multiplying it with $\boldsymbol{M}^{-1}$. From the traces of the rescaled covariance matrices, we do not observe significant differences across the methods, although the variance is consistently smaller for smaller values of $\sigma_{\text{log}}$, as expected. In terms of computational cost, we note that the hyperparameter tuning is performed offline only once and is therefore independent of the value of $\sigma_\text{log}$. Consequently, the values in the Optuna column are identical across all tables. 
Finally, the computational cost for Methods B through E is also in line with Method A (which has the lowest cost due to it being an analytical function), even despite the addition of a low-fidelity function evaluation (Methods C and E). However, Method F has a higher cost due to the complexity of the underlying RealNVP normalizing flow architecture.}

\begin{table}[!ht]
\centering
\caption{\chloerev{Accuracy, variance, and computational cost of the posterior distributions obtained using $N=100$ samples, for the borehole function test case with \textit{uniform prior}. Results are averaged over 10 independent experiments.}}
\begin{tabular}{c c c c c} \toprule \multirow{2}{*}{\bf Method} & \multirow{2}{*}{\bf KL divergence} & \multirow{2}{*}{tr($M^{-1}\mathbb{C}^{\mu}(\cdot)M^{-1}$)} & \multicolumn{2}{c}{\bf Cost}\\ & & & {\bf Optuna} & {\bf Posterior}\\ \midrule {\bf A} & 0.0 & 6.01 & - & 126 s\\
{\bf B} & 0.0722 $\pm$ 0.01739 & $6.42 \pm 0.50$ & $514 \pm 114$ s & $136 \pm 3$ s\\
{\bf C} & $\mathbf{0.0575 \pm 0.0223}$ & $6.16 \pm 0.40$ & $547 \pm 109$ s & $140 \pm 3$ s\\
{\bf D} & 0.0832 $\pm$ 0.0155 & $6.21 \pm 0.37$ & $2058 \pm 242 $ s & $137 \pm 2$ s\\ 
{\bf E} & 0.0651 $\pm$ 0.0154 & $6.23 \pm 0.39$ & $1836 \pm 323$ s & $141 \pm 2$ s\\ 
{\bf F} & $0.0591 \pm 0.0274$ & $6.23 \pm 0.15$ & $4129 \pm 768$ s & $4141 \pm 1572$ s\\ 
\bottomrule \end{tabular}
\label{tab:borehole_N_100_uniform}
\end{table}

\begin{table}[!ht]
\centering
\caption{\chloerev{Accuracy, variance, and computational cost of the posterior distributions obtained using $N=100$ samples, for the borehole function test case with \textit{truncated normal distribution} with $\sigma_{\text{log}}=0.5$. Results are averaged over 10 independent experiments.}}
\begin{tabular}{c c c c c} \toprule \multirow{2}{*}{\bf Method} & \multirow{2}{*}{\bf KL divergence} & \multirow{2}{*}{tr($M^{-1}\mathbb{C}^{\mu}(\cdot)M^{-1}$)} & \multicolumn{2}{c}{\bf Cost}\\ & & & {\bf Optuna} & {\bf Posterior}\\ \midrule {\bf A} & 0.0 & 3.13 & - & 133 s\\
{\bf B} & $0.0338 \pm 0.0114$ & $3.32 \pm 0.28$ & $514 \pm 114$ s & $138\pm23$ s\\
{\bf C} & $0.0347 \pm 0.0027$ & $3.17 \pm 0.21$ & $547 \pm 109$ s & $149 \pm 2$ s\\
{\bf D} & $0.0381 \pm 0.0018$ & $3.36 \pm 0.23$ &  $2058 \pm 242 $ s & $216 \pm 179$ s\\ 
{\bf E} & $0.0339 \pm 0.0025$ & $3.34 \pm 0.24$ & $1836 \pm 323$ s & $150 \pm 3$ s\\ 
{\bf F} & $\mathbf{0.0282 \pm 0.0143}$ & $3.36 \pm 0.19$ & $4129 \pm 768$ s & $4042 \pm 1892$ s\\ 
\bottomrule \end{tabular}
\label{tab:borehole_N_100_gaussian_0p5}
\end{table}

\begin{table}[!ht]
\centering
\caption{\chloerev{Accuracy, variance, and computational cost of the posterior distributions obtained using $N=100$ samples, for the borehole function test case with \textit{truncated normal distribution} with $\sigma_{\text{log}}=0.1$. Results are averaged over 10 independent experiments.}}
\begin{tabular}{c c c c c} \toprule \multirow{2}{*}{\bf Method} & \multirow{2}{*}{\bf KL divergence} & \multirow{2}{*}{tr($M^{-1}\mathbb{C}^{\mu}(\cdot)M^{-1}$)} & \multicolumn{2}{c}{\bf Cost}\\ & & & {\bf Optuna} & {\bf Posterior}\\ \midrule {\bf A} & 0.0 & 0.1446 & - & 133 s\\
{\bf B} & $0.0470 \pm 0.0019$ & $0.1445 \pm 0.0037$ &  $514 \pm 114$ s & $145 \pm 3$ s\\
{\bf C} & $0.0454 \pm 0.0021$ & $0.1422 \pm 0.0037$ & $547 \pm 109$ s & $148 \pm 3$ s\\
{\bf D} & $0.0496 \pm 0.0034$ & $0.1424 \pm 0.0049$ &  $2058 \pm 242 $ s & $145 \pm 2$ s\\ 
{\bf E} & $0.0468 \pm 0.0017$ & $0.1410 \pm 0.0030$ & $1836 \pm 323$ s & $150 \pm 3$ s\\ 
{\bf F} & $\mathbf{0.0326 \pm 0.0214}$ & $0.1414 \pm 0.0045$ & $4129 \pm 768$ s & $3906 \pm 1750$ s\\ 
\bottomrule \end{tabular}
\label{tab:borehole_N_100_gaussian_0p1}
\end{table}

\begin{table}[!ht]
\centering
\caption{\chloerev{Accuracy, variance, and computational cost of the posterior distributions obtained using $N=100$ samples, for the borehole function test case with \textit{truncated normal distribution} with $\sigma_{\text{log}}=0.01$. Results are averaged over 10 independent experiments.}}
\begin{tabular}{c c c c c} \toprule \multirow{2}{*}{\bf Method} & \multirow{2}{*}{\bf KL divergence} & \multirow{2}{*}{tr($M^{-1}\mathbb{C}^{\mu}(\cdot)M^{-1}$)} & \multicolumn{2}{c}{\bf Cost}\\ & & & {\bf Optuna} & {\bf Posterior}\\ \midrule {\bf A} & 0.0 & 0.00364 & - & 67 s\\
{\bf B} & $0.0582 \pm 0.0117$ & $0.00373 \pm 0.00005$ &  $514 \pm 114$ s & $72 \pm 1$ s\\
{\bf C} & $0.0540 \pm 0.0122$ & $0.00370 \pm 0.00004$ & $547 \pm 109$ s & $75 \pm 1$ s\\
{\bf D} & $0.0576 \pm 0.0140$ & $0.00368 \pm 0.00006$ & $547 \pm 109$ s & $73 \pm 1$ s\\ 
{\bf E} & $\mathbf{0.0511 \pm 0.0145}$ & $0.00370 \pm 0.00004$ &  $2058 \pm 242 $ s &  $75 \pm 1$ s\\ 
{\bf F} & $0.0588 \pm 0.0059$ & $0.00371 \pm 0.00003$ &$4129 \pm 768$ s & $2119 \pm 720$s\\ 
\bottomrule \end{tabular}
\label{tab:borehole_N_100_gaussian_0p01}
\end{table}

\section{Cardiovascular simulations} \label{sec:cardio}

We now compare the effectiveness of Methods B to F to reduce the computational cost of parameter estimation in challenging scenarios involving realistically expensive high-fidelity cardiovascular models.
In particular, we consider a three-element Windkessel ODE model and a patient-specific model of the aorto-iliac bifurcation.

\subsection{Zero-dimensional cardiovascular model} \label{sec:circuits}

In this section we infer the parameters of a three-element Windkessel (RCR) model from the solution of inverse problems computed from single-fidelity or bi-fidelity surrogates.
Both high- and low-fidelity models are lumped parameter networks (LPNs) -- or circuit models, or zero-dimensional hemodynamic models -- that capture the relationship between blood flow, pressure, and inertia and resistance and compliance of vessels in the cardiovascular system. 
Due to their negligible computational cost, they are often used as boundary conditions for more realistic one- and three-dimensional formulations.

The three-parameter (RCR) Windkessel model is governed by the following set of differential equations
\begin{equation}
    Q_p = \frac{P_p - P_c}{R_p},\quad Q_d = \frac{P_c - P_d}{R_d},\quad \frac{dP_c}{dt} = \frac{Q_p - Q_d}{C},
\end{equation}
while the two-parameter (RC) Windkessel model is governed by
\begin{equation}
    Q_p = \frac{P_p - P_d}{R}, \quad\frac{dP_p}{dt} = \frac{Q_p - Q_d}{C}.
\end{equation}

\begin{figure}[!htb]
    \centering
    \begin{subfigure}[m]{0.38\textwidth}
    \begin{overpic}[abs,unit=1mm,width=0.8\textwidth]{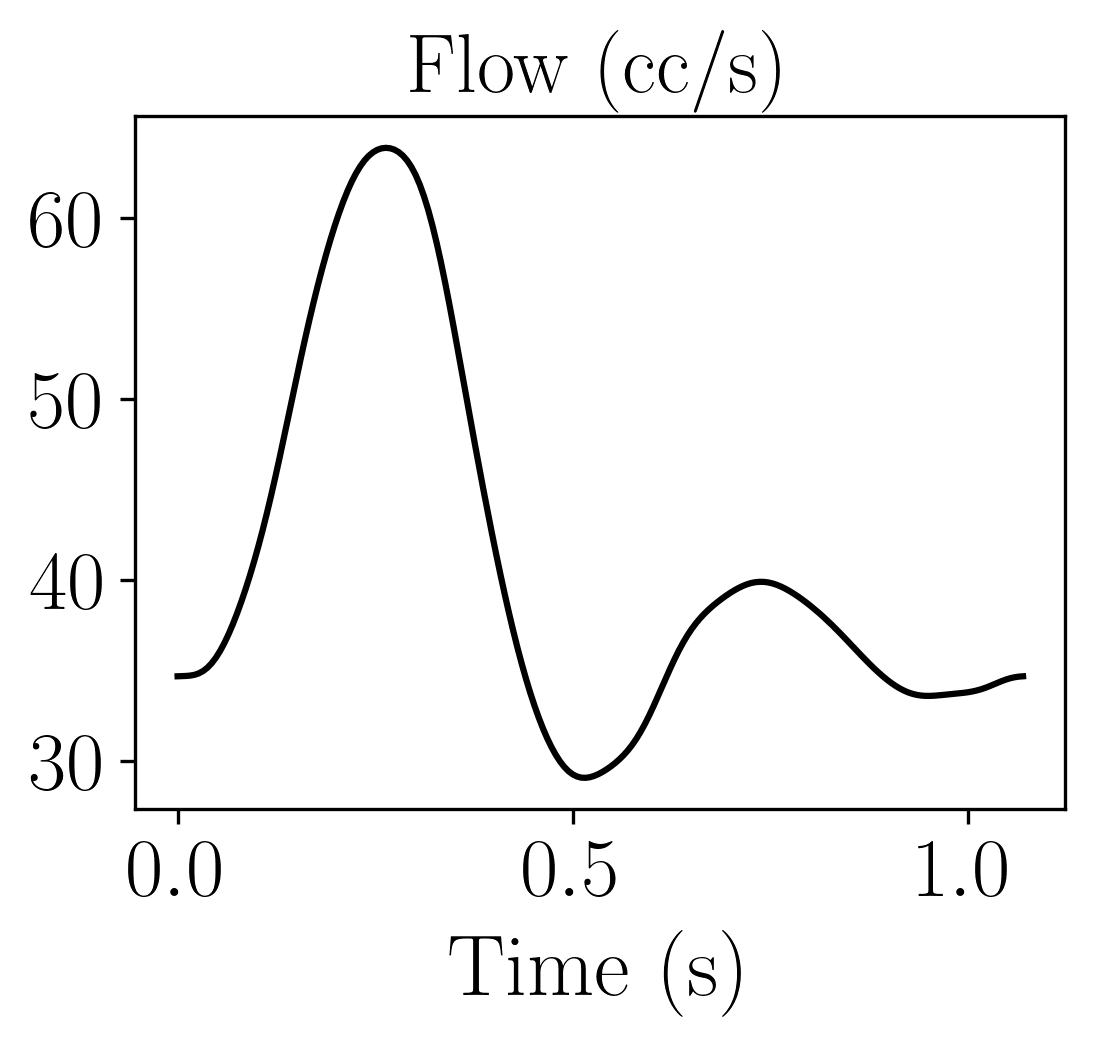}
    \end{overpic}
    \caption{Time history of the proximal flow rate $Q_p$.}
    \end{subfigure}
    \begin{subfigure}[m]{0.58\textwidth}
    \begin{subfigure}{\textwidth}
    \begin{overpic}[abs,unit=1mm,width=0.9\textwidth]{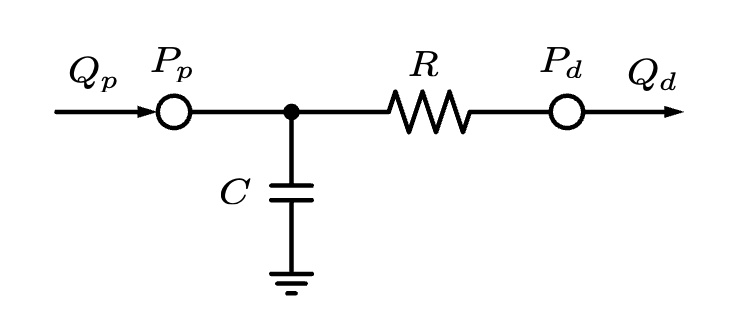}
    \end{overpic}
    \caption{Two-element Windkessel (RC) model.}
    \end{subfigure}
    
    \vspace{20pt}
    
    \begin{subfigure}{\textwidth}
    \centering
    \begin{overpic}[abs,unit=1mm,width=\textwidth]{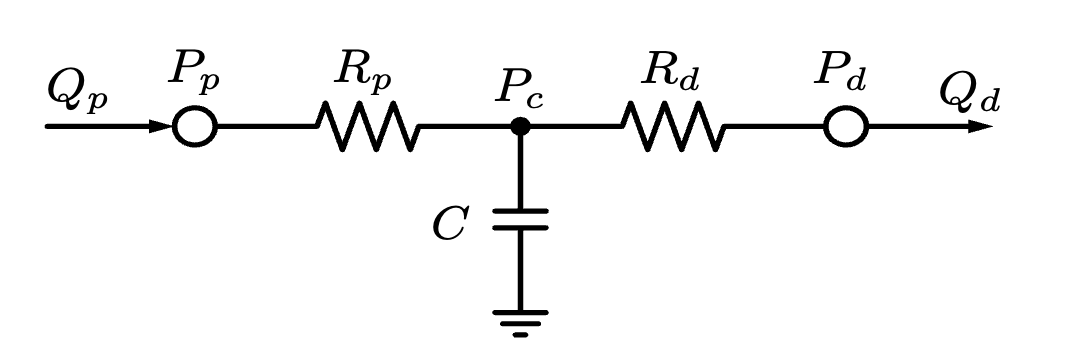}
    \end{overpic}
    \caption{Three-element Windkessel (RCR) model.}
    \end{subfigure}
    \end{subfigure}
    \caption{Inflow profile and circuit diagrams of the RC and RCR models.}\label{fig:circuit_inflow_and_diagrams}
\end{figure}

\cref{fig:circuit_inflow_and_diagrams} shows the periodic time history of the abdominal aortic inflow~\cite{AA_inflow} as well as the circuit diagrams of the RC and RCR models. 

We integrate both initial value problems in time using an implicit fifth-order Radau scheme from the \texttt{solve\_ivp} function, provided by the \texttt{scipy.integrate} package. 
A prescribed inlet condition is assigned through the time history of $Q_p$, as shown in~\cref{fig:circuit_inflow_and_diagrams}.
Additionally, we set the prior parameter bounds as $R_p \in [500, 1500]$ Barye $\cdot$ s / mL, $R_d \in [500, 1500]$ Barye $\cdot$ s / mL, and $C \in [10^{-5}, 10^{-4}]$ mL/Barye, and set the distal pressure to be constant in time and equal to 55 mmHg.
We finally consider a cardiac cycle with duration 1.07 s, and run each simulation for 10 cycles.

The \emph{true} parameters values are assumed equal to $R_p = R_d = 1000$ Barye $\cdot$ s/mL and $C = 5 \cdot 10^{-5}$ mL/Barye, leading to a proximal pressure with $P_{\text{min}} = 100.96$ mmHg, $P_{\text{max}} = 148.02$ mmHg, and $P_{\text{avg}} = 116.50$ mmHg.
Synthetic observations are generated from $\mathcal{N}(\boldsymbol{\mu}, \boldsymbol{\Sigma})$, where $\boldsymbol{\Sigma}$ is a diagonal matrix with entries $(5.05, 7.40, 5.83)^T$. We use fifty observations for the analysis.
We construct a shared parameter space, $(R_p, R_d, C)$, for both models to calculate the discrepancy between the models. 
In doing so, we assume $R=R_p+R_d$ for the low-fidelity model. 
To construct the posterior distribution, we hold constant the prior distribution and alter the likelihood as before for the different methods. 
Posterior samples are generated using differential evolution adaptive metropolis (DREAM~\cite{vrugt_2009}), which combines differential evolution~\cite{DE_1997} with self-adaptive randomized subspace sampling. 
Specifically, we use the \texttt{PyDream} Python package associated with the paper (\url{https://github.com/LoLab-MSM/PyDREAM/tree/master}).
We use a uniform distribution that samples from the log space of the bounds mentioned previously. 
We run five chains in parallel for 50,000 iterations and reach Gelman--Rubin statistics that are below 1.01 for each of the methods discussed in Section~\ref{sec:surrogates} and~\ref{sec:model_error}.

We again create a \chloerev{training} data set of size $N=100$.
For each method, we evaluate 10 different experiments and report the mean and standard deviation of the Kullback--Leibler (KL) divergences calculated using k-nearest neighbors, as suggested in~\cite{KLD_perez_2008}. 
In~\cref{tab:circuit_N_100}, we see that the KL divergence is comparable for all five methods (B to F).
The table also shows a larger trace of the covariance matrix for method F, which is expected since the distribution of the model discrepancy and noise are jointly considered, and approximated by normalizing flow.

\begin{table}[!ht]
\centering
\caption{Accuracy, variance, and computational cost of the posterior distributions obtained using $N=100$ samples, for the three-element Windkessel test case. Results are averaged over 10 independent experiments. Note that the third column reports on the mean only.}
\resizebox{\textwidth}{!}{
\begin{tabular}{c c c c c c}
\toprule
\multirow{2}{*}{\bf Method} & \multirow{2}{*}{\bf KL Divergence} & \multirow{2}{*}{diag($\mathbb{C}^{\mu}(\cdot)$)} & \multicolumn{2}{c}{\bf Cost}\\
& & & {\bf Optuna} & {\bf Posterior}\\
\midrule
{\bf A} &  0.0 & [4.91e4, 4.98e4, 3.15e-10] & - & 5453 s\\
{\bf B} &  $0.0161 \pm 0.0006$ & [5.63e4, 5.71e4, 3.48e-10] & $\bm{394 \pm 55}$ s & $\bm{188 \pm 8}$ s\\
{\bf C} &  $0.0158 \pm 0.0005$ & \textbf{[5.15e4, 5.23e4, 3.18e-10]} & $486 \pm 78$ s & $3894 \pm 294$ s\\
{\bf D} & $0.0170 \pm 0.0008$ & [6.66e4, 6.75e4, 4.17e-10] & $1483 \pm 76 $ s & $188 \pm 11$ s\\
{\bf E} & $\boldsymbol{0.0158 \pm  0.0006}$ & [5.22e4, 5.29e4, 3.27e-10] & $1531 \pm 94$ s & $3791 \pm 280$ s\\
{\bf F} & $0.0188 \pm 0.0040$ & [5.27e4, 5.35e4, 3.33e-10] & $3880 \pm 372$ s & $3246 \pm 354$ s\\
\bottomrule
\end{tabular}}
\label{tab:circuit_N_100}
\end{table}

\begin{figure}[!htb]
    \centering
    \includegraphics[width=0.9\linewidth]{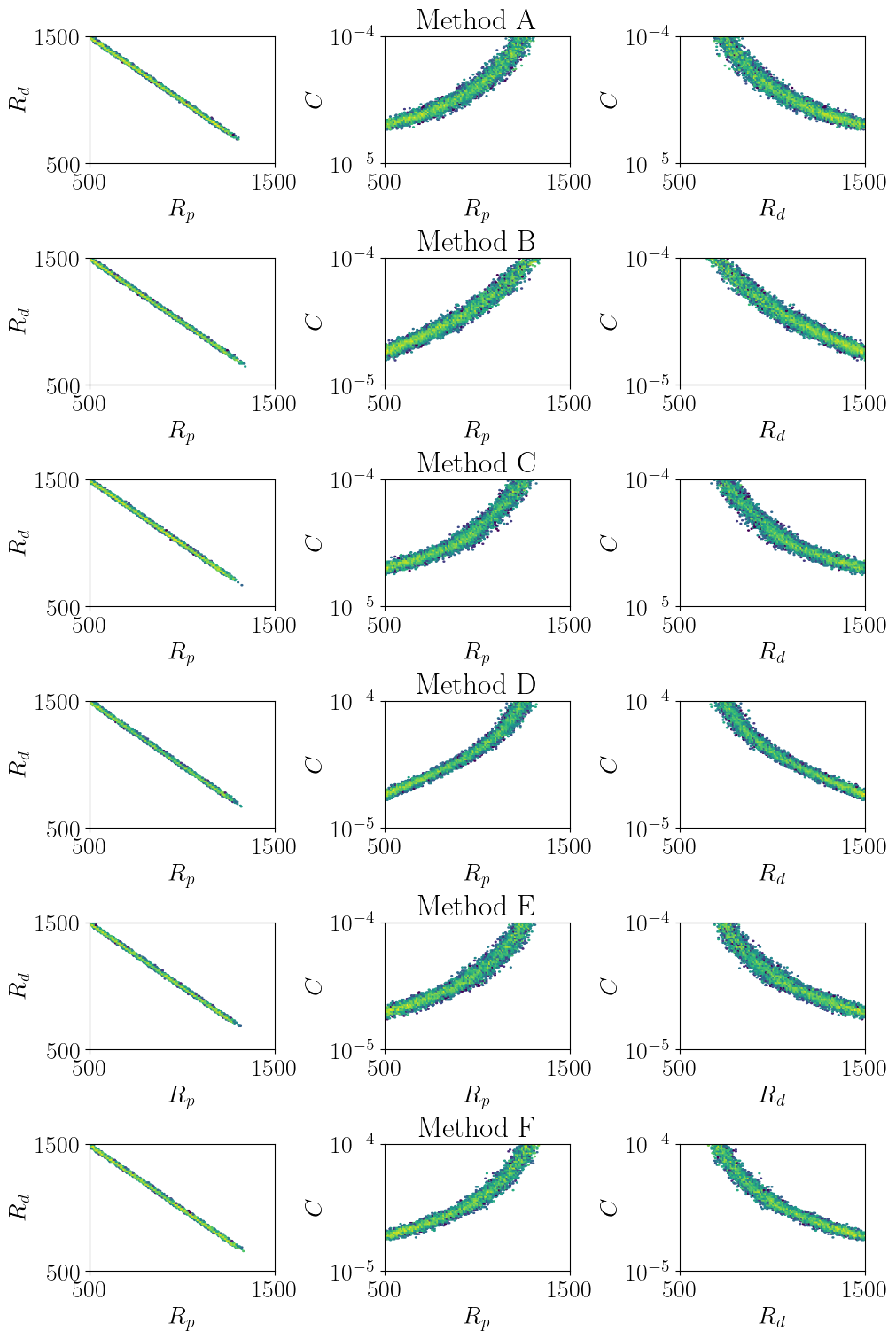}
    \caption{Samples from the posterior distributions given by the six different methods using $N=100$ data, for the circuit example.}
    \label{fig:circuit_posteriors}
\end{figure}

We display the results in~\cref{fig:circuit_posteriors}. 
We remark on the physical significance of the posterior distributions. 
First, there is negative linear correlation between $R_d$ and $R_p$, as the mean pressure depends only on the sum $R_p+R_d$, so if the mean pressure is fixed, then an increase in $R_p$ must be compensated by reducing $R_d$. 
Second, there is nonlinear positive correlation between $R_p$ and $C$; as $R_p$ increases, then $R_d$ decreases, so the flow into the capacitor increases, and $C$ must be increased to leave the pulse pressure unaltered. 
This argument also explains the negative nonlinear correlation between $R_d$ and $C$. 
Note also the lack of identifiability clearly visible from the resulting posterior distributions, where multiple MAP estimates lay on a one-dimensional manifold. 
Note further that \texttt{PyDREAM} sampler uses \emph{reflective} boundary conditions, where points lying outside the uniform sample space will reflect back inside. 
This leads to a tendency to overestimate the sample density on the boundary.

\subsection{Aorto-iliac bifurcation}

The application discussed in this section leverages two model formulations that are widely used in vascular hemodynamics, i.e., three- and zero-dimensional approximations.
Note that, as the dimensionality of the model formulation increases, so does the computational cost and the solution accuracy.
Transitioning between model formulations is also particularly convenient, as both approaches share a common model creation pipeline which begins with acquiring MRI or CT medical images, creating vessel centerline, performing the segmentation of the vessel cross-section, lofting and solid model creation, meshing and applying initial and boundary conditions~\cite{simvascular}.
At the end of this process the user can choose to solve various equations over the discretized domain $\boldsymbol{\Omega}^{3D}\subset\R^{3}$. 
The evolution of the three-dimensional velocity and pressure field can be represented by the Navier-Stokes equations in strong form~\cite{1967_batchelor} as 
\begin{equation} \label{eq:3d_nse_cont}
\begin{aligned}
    \rho(\dot{\boldsymbol{u}} + \boldsymbol{u} \cdot \nabla \boldsymbol{u}) &= \nabla \cdot \boldsymbol{\tau} + \rho \boldsymbol{b}, &&\boldsymbol{x} \in \boldsymbol{\Omega}^{3D}, \quad t > 0 \\
    \boldsymbol{\tau} &= -p \boldsymbol{I} + \mu (\nabla \boldsymbol{u} + \nabla \boldsymbol{u}^{\intercal}), &&\boldsymbol{x} \in \boldsymbol{\Omega}^{3D}, \quad t > 0 \\
    \nabla \cdot \boldsymbol{u} &= 0, &&\boldsymbol{x} \in \boldsymbol{\Omega}^{3D}, \quad t > 0.
\end{aligned}
\end{equation}
These equations can be expressed in cylindrical coordinates, further simplified by assuming an axisymmetric velocity profile, and integrated over the cross section yielding the one-dimensional Navier-Stokes equations~\cite{1973_hughes, 1974_hughes}
\begin{equation} \label{eq:1d_momentum_continuity}
\begin{aligned}
    \frac{\partial Q}{\partial t} + \frac{4}{3} \frac{\partial}{\partial z} \Big( \frac{Q^2}{S} \Big) + \frac{S}{\rho} \frac{\partial p}{\partial z} &= Sf - N \frac{Q}{S} + \frac{\mu}{\rho} \frac{\partial^2 Q}{\partial z^2}, &&z \in \boldsymbol{\Omega}^{1D}, \quad t > 0, \\
    \frac{\partial S}{\partial t} + \frac{\partial Q}{\partial z} &= 0, &&z \in \boldsymbol{\Omega}^{1D}, \quad t > 0,
\end{aligned}
\end{equation}
where a third constitutive equation $S=S(p)$ is typically used as \emph{closure}, and for a parabolic velocity profile (Poiseuille flow) one has $N = 8 \pi \mu / \rho$.

By further linearization of the above equations around rest conditions~\cite{rideout2008difference}, one obtains new equations that are formally identical to those governing the evolution of voltage and current in an electrical circuit (the so-called \emph{hydrodynamic analogy}), leading to
\begin{equation}
\begin{aligned}
    C \frac{\partial p_1}{\partial t} + Q_2 - Q_1 &= 0, &&t>0 \\
    L \frac{\partial Q_2}{\partial t} + P_2 - P_1 &= -RQ_2, &&t>0,
\end{aligned}
\end{equation}
which are equivalent to the RLC equations
\begin{equation}
\begin{aligned}
    \Delta Q = C \dot{P}, \quad \Delta P = RQ, \quad \Delta P = L \dot{Q},
\end{aligned}
\end{equation}
as we discussed earlier in \cref{sec:circuits}. 
Under Poiseuille flow conditions, and ideal cylindrical vessel walls modeled as a thin, linear elastic, and homogeneous shell, the RLC constants are approximated by 
\begin{equation}
    R=\frac{8 \mu l}{\pi r^4}, \quad C = \frac{3 l \pi r^3}{2 E h}, \quad L = \frac{\rho l}{\pi r^2},
\end{equation}
where $l$ the vessel length, $E$ the Young's modulus of the vascular tissue, and $r$ the lumen radius. Note that for rigid-wall simulations, capacitance is set to zero.
In addition, the blood density and viscosity used in the fluid simulations are assumed equal to $\rho = 1.06$ g/cm$^3$ and $\mu = 0.04$ g/cm/s, respectively.

\begin{figure}[!htb]
    \centering
    \begin{subfigure}[m]{0.38\textwidth}
    \begin{overpic}[abs,unit=1mm,width=0.85\textwidth]{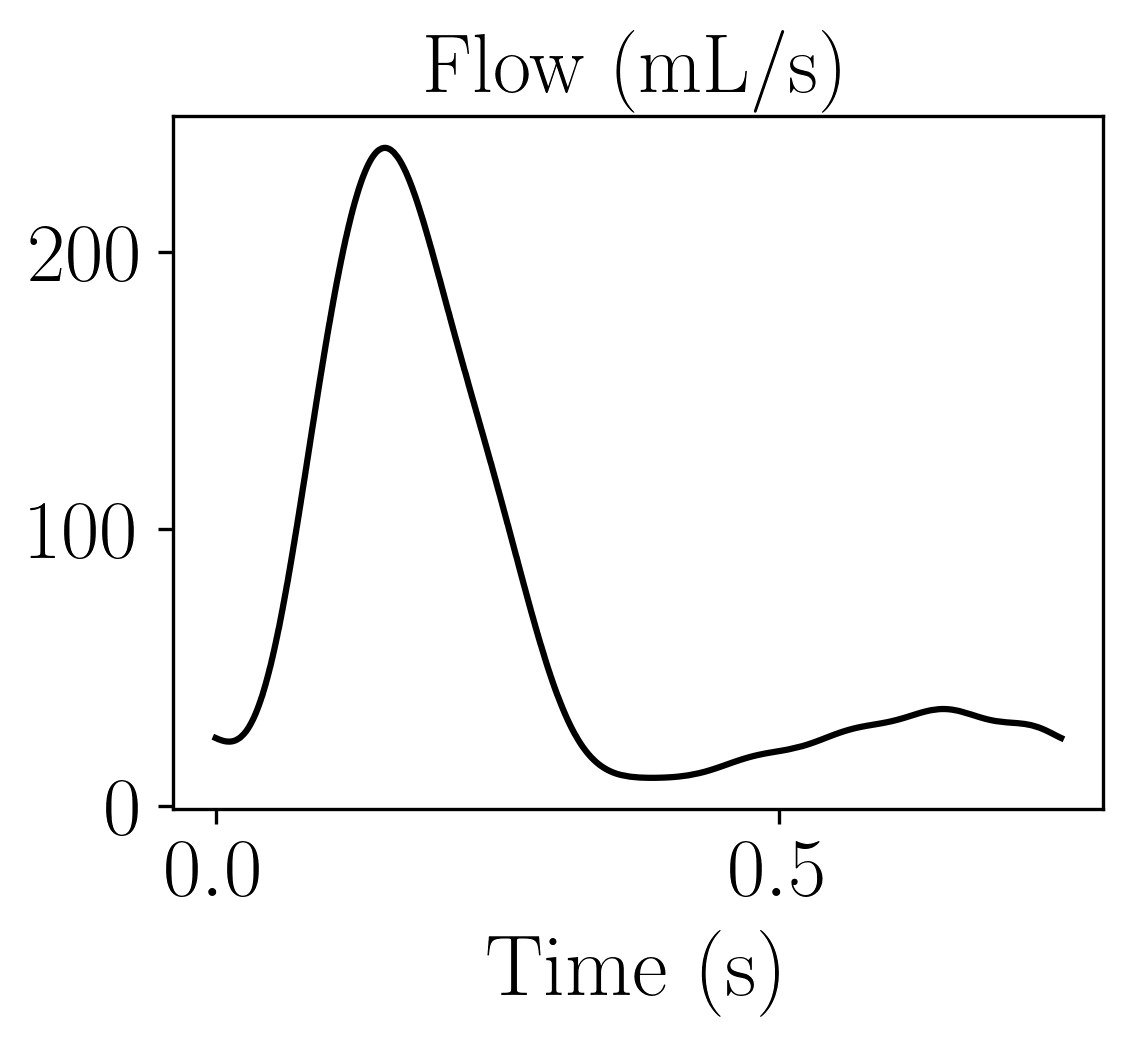}
    \end{overpic}
    \caption{Selected inflow at abdominal aorta.}
    \end{subfigure}
    \begin{subfigure}[m]{0.58\textwidth}
    \begin{subfigure}{\textwidth}
    \begin{overpic}[abs,unit=1mm,width=\textwidth]{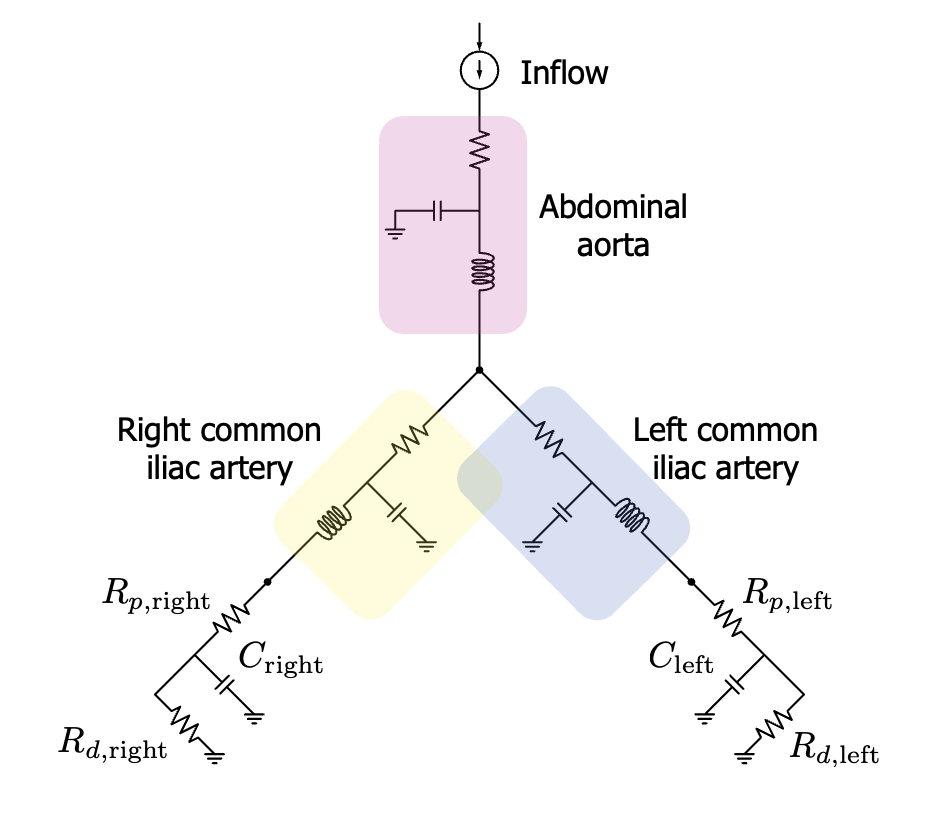}
    \end{overpic}
    \caption{Low-fidelity zero-dimensional approximation.}
    \end{subfigure}
    
    \vspace{20pt}
    
    \begin{subfigure}{\textwidth}
    \centering
    \begin{overpic}[abs,unit=1mm,width=\textwidth]{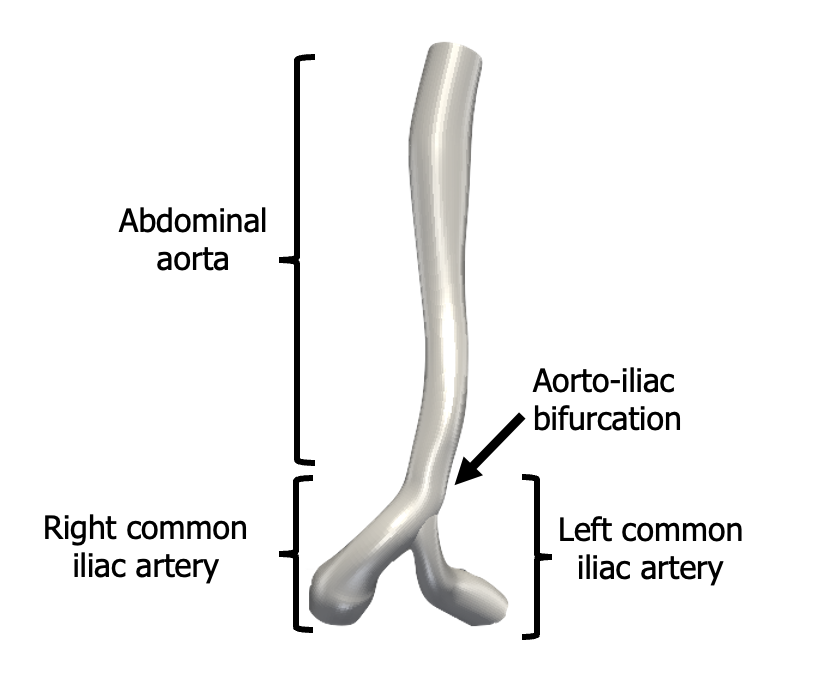}
    \end{overpic}
    \caption{High-fidelity three-dimensional model.}
    \end{subfigure}
    \end{subfigure}
    \caption{Inflow time history and high- and low-fidelity models of the aorto-iliac bifurcation test case}\label{fig:aobif_model_setup}
\end{figure}

We use a patient-specific aorto-iliac bifurcation model that is accessible from the \texttt{SimVascular} open source project via the SimTK platform \cite{simvascular}. 
Also, note that this model is a truncated version of patient ``0030\_H\_ABAO\_H'' from the Vascular Model Repository~\cite{vmr} (\url{www.vascularmodel.com/}). 
We impose a patient-specific inflow waveform with typical values for flow rate at the abdominal aorta.
Figure~\ref{fig:aobif_model_setup} shows a rendering of the selected anatomy, its zero-dimensional analog, and the prescribed patient-specific inflow.

We assign three-element Windkessel boundary conditions to each of the two outlets. 
We then tune a zero-dimensional approximation with the same boundary conditions using Nelder-Mead optimization to reach target pressures equal to 120/80 mmHg, representative of a healthy subject. 
We determine the total peripheral resistance from the mean inflow and the mean cuff BP ($P_{\text{mean}}$), with $P_{\text{mean}}$ defined as
\begin{equation}
    P_{\text{mean}} = \frac{1}{3} P_{\text{sys}} + \frac{2}{3} P_{\text{dia}},
\end{equation}
where the systolic pressure $P_{\text{sys}}$ is 120 mmHg and diastolic pressure $P_{\text{dia}}$ is 80 mmHg. The noise is given by $\bm{\Sigma} = \text{diag}(0.05 P_{\text{dia}}, 0.05 P_{\text{sys}}, 0.05 P_{\text{mean}})$.
The total peripheral resistance is then divided among the branches inversely proportional to the cross-sectional outlet area of each branch, or, in other words, we distribute the peripheral resistance using Murray's law with exponent equal to 2~\cite{sherman1981connecting}.
We also assume a proximal-to-distal resistance ratio of 1/10 for each outlet, following~\cite{MAHER21_geometric}. The total capacitance is likewise divided between the branches in proportion to the cross-sectional outlet area of each branch. 
The specific baseline values for each outlet are listed in ~\cref{tab:aobif_baseline_vals}.

\begin{table}[!ht]
    \centering
    \caption{True RCR boundary condition parameters}
    \begin{tabular}{l c c c}
    \toprule
    {\bf Vessel} & {\bf $R_p$ [$\text{dynes}\cdot\text{s}/\text{cm}^5$]} & {\bf $C$ [cm$^5$/dyne]} & {\bf $R_d$ [$\text{dynes}\cdot\text{s}/\text{cm}^5$]}\\
    \midrule
     {\bf Right iliac} & 394 & 2.30 $\times 10^{-4}$ & 3844 \\
     {\bf Left iliac} & 306 & 2.96 $\times 10^{-4}$ & 2986 \\
         \bottomrule
    \end{tabular}
    \label{tab:aobif_baseline_vals}
\end{table}

We vary the total baseline resistance and capacitance by a standard deviation of 30\% of their nominal values. 
Using the same parameters, we run 100 rigid-wall simulations of the three-dimensional model using \texttt{svMultiPhysics}, an open-source, parallel, finite element multi-physics solver, which is a part of the SimVascular open source platform, and 100 simulations of the zero-dimensional low-fidelity model using \texttt{svZeroDSolver}, where we set $C=0$ for rigid wall behavior. 
We use 75 of the $N=100$ simulations for the training data set, and the remaining 25 as the validation set. 
Each three-dimensional simulation took up to 17 hours to evaluate on 128 processors on AMD EPYC 7742 on the Expanse machine at the San Diego Supercomputing Cluster. 
We remark that, due to the computationally expensive nature of these cardiovascular simulations, it is not possible to construct the entire posterior distribution for the high-fidelity model for baseline comparison. 

\begin{figure}[!htb]
    \centering
    \includegraphics[width=0.8\linewidth]{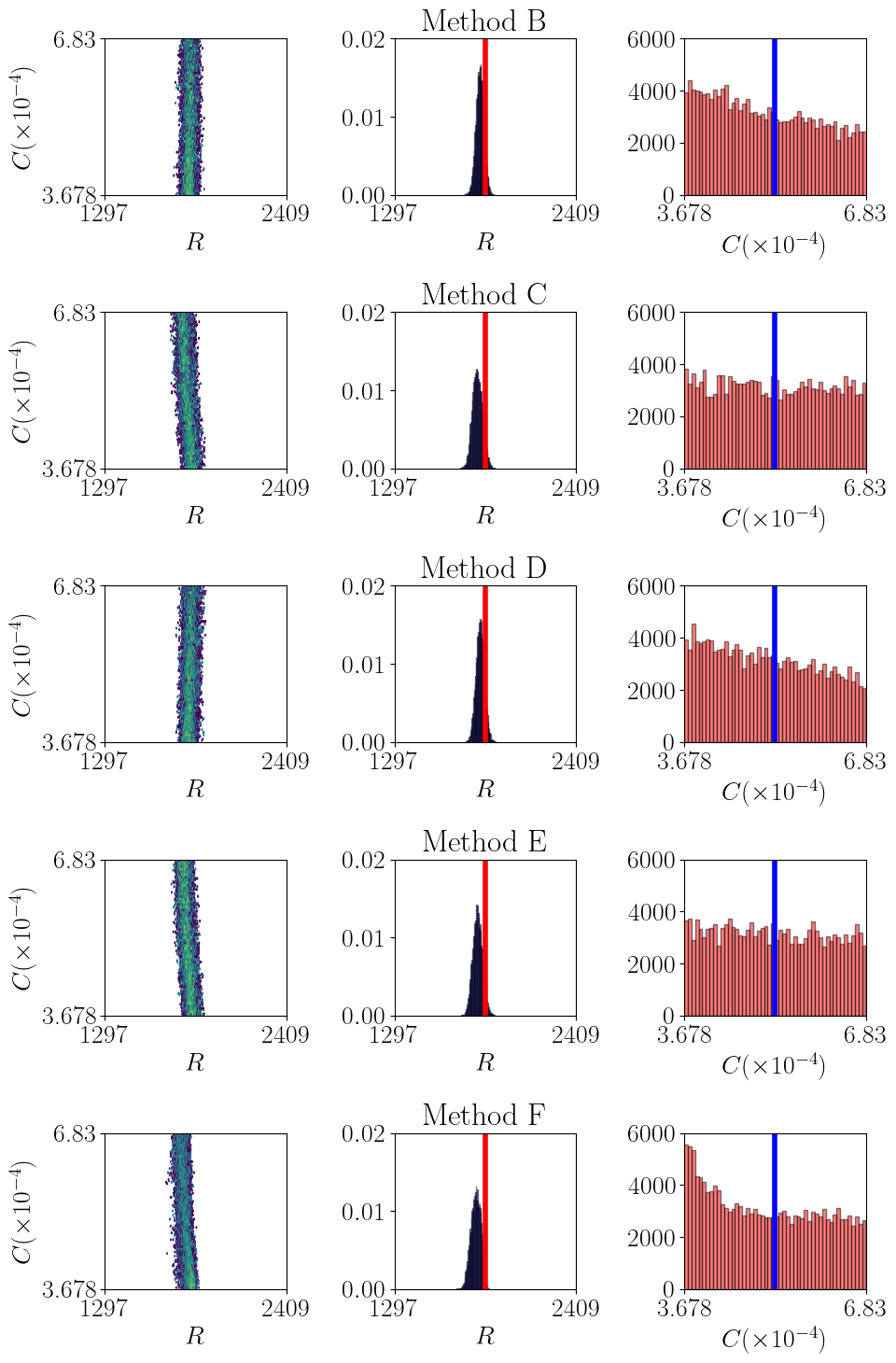}
    \caption{Left: Samples from the posterior distributions given by the five different methods using $N=100$ data for the aorto-iliac bifurcation example. Center and right: Samples from the marginal distributions of $R$ and $C$ given by the five different methods using $N=100$ data. True nominal values are represented by a red and blue vertical line, for the resistance and capacitance, respectively.}
    \label{fig:aobif_posteriors}
\end{figure}

We run five parallel chains for 10,000 iterations and reach Gelman--Rubin statistics that are below 1.01 for each of the methodologies. 
The results for $N=100$ are shown in \cref{fig:aobif_posteriors}. We observe that while the total resistance is identifiable, the capacitance exhibits a practically flat marginal posterior, which is typical of unimportant parameters. 

%
In \cref{tab:aobif_N_100}, we also compare the computational time of each methodology. Note here that we report on the results of one experiment, rather than taking the mean and standard deviation of 10 different initializations of the training data set, due to the computationally expensive nature of high-fidelity three-dimensional cardiovascular simulations.
We note that hyperparameter tuning takes longest for Method F, as NeurAM and NF are tuned independently. Method B and C take the least amount of time, as they are simple fully connected neural networks without dimensionality reduction. Notice, further, that Methods B and D take the least amount of time to compute the posterior, as only the surrogate is evaluated for the Gaussian likelihood calculation. In contrast, the discrepancy-based approaches take longer  because the low-fidelity model is computed in addition to each surrogate evaluation at each posterior sample. Note, further, that the posterior evaluation times for the discrepancy-based approaches in the aorto-iliac bifurcation example are less than those of the zero-dimensional model in \cref{sec:circuits}. This is because the low-fidelity model used in this example is svZeroDSolver, whereas the one in the previous example solves the initial value problem with an implicit fifth-order Radau scheme, and also because this example requires a smaller number of iterations per chain to reach GR convergence, as the input dimension is smaller.
Finally, the variance for Method F is larger than Method E, as expected.
%

\begin{table}[!ht]
\centering
\caption{Accuracy, variance, and computational cost of the posterior distributions obtained using $N=100$ samples, for the aorto-iliac bifurcation example. Results are reported for one single experiment.}
\begin{tabular}{c c c c}
\toprule
\multirow{2}{*}{\bf Method} & \multirow{2}{*}{diag($\mathbb{C}^{\mu}(\cdot)$)} & \multicolumn{2}{c}{\bf Cost}\\
& & {\bf Optuna} & {\bf Posterior}\\
\midrule
{\bf A} & - & - & -\\
{\bf B} & [587, 1.77e-4] & 661 s & 47 s\\
{\bf C} & [941, 8.41e-9] & 395 s & 450 s\\
{\bf D} & [668, 4.35e-4] & 2476 s & 47 s\\
{\bf E} & [868, 8.32e-9] & 2606 s & 449 s\\
{\bf F} & [890, 8.91e-9] & 4115 s & 340 s\\
\bottomrule
\end{tabular}
\label{tab:aobif_N_100}
\end{table}

\section{Conclusion}\label{sec:conclusion}

This study compared the numerical efficiency of various strategies for the solution of inverse problems with computationally expensive models. 
In particular, we considered combinations of adaptive MCMC sampling with surrogate models trained from single- and multi-fidelity data sets.
We considered a broad range of surrogates from neural networks that directly approximate a high-fidelity quantity of interest to multi-fidelity discrepancy-based formulations leveraging a computationally inexpensive low-fidelity approximant. 
We also analyzed the performance of reduced-order models generated through nonlinear dimensionality reduction (NeurAM~\cite{ZGSMD24}), based on either a high-fidelity model or the discrepancy between such a model and low-fidelity approximation. 
Finally, we employed a new approach where a non-Gaussian likelihood combining model discrepancy and observational noise was estimated using normalizing flow. 

Our performance assessment is based on \chloerev{five} numerical examples, including \chloerev{three} closed-form maps and two applications involving boundary condition tuning under uncertainty in computational hemodynamics. 
In the above test cases, all the methods provided satisfactory results and were able to successfully capture the features of the posterior distribution, including ridges, sharply peaked distributions, strong linear and nonlinear correlations, and unimportant parameters with flat marginal posterior. 
Differences in the Hellinger distance (closed-form maps) and KL divergence (hemodynamic models) suggest that a discrepancy formulation is generally preferable when computationally inexpensive low-fidelity models are available and the discrepancy is smoother than the high-fidelity function. 
We examined the variances of the posterior grids and samples, respectively, by taking the trace of the closed-form solutions and assessing each diagonal element for the cardiovascular applications. We consistently observed that normalizing flow overestimated the variance for all the methods. 
Training times including hyperparameter optimization ranged from minutes for neural network surrogates to tens of minutes for NeurAM based surrogates and normalizing flow, and were mainly affected by the number of parameters in the selected architecture.
The approach of estimating the density of both measurement noise and model discrepancy results in posterior distributions that are aligned with those produced by other methods, and where variance overestimation remains limited. The only downside of this approach appears to be a slight overhead in evaluating posterior samples sequentially, due to the number of parameters in the normalizing flow architecture, which includes one neural network per RealNVP transformation.

Future work will focus on boundary condition tuning under uncertainty for models with a larger number of outlets, resulting in higher dimensional inverse problems. 
We will also work toward incorporating the methods discussed in this study in a more general framework for boundary condition tuning in cardiovascular modeling. 
\chloerev{We further note that there is potential to extend the proposed framework to real experimental data. In such a case, the experimental data could serve as the high-fidelity data, simulation data as the next level of low-fidelity data, and the 0D model as the lowest fidelity source.}

\subsection*{Acknowledgments}

This work is supported by the Yansouni Family Stanford Graduate Fellowship (CHC) and NSF grant \#2105345 (ALM). 
The authors also acknowledge support from NSF CAREER award \#1942662 and a NSF CDS\&E award \#2104831 (DES), and from NIH grant \#1R01HL167516 {\it Uncertainty aware virtual treatment planning for peripheral pulmonary artery stenosis} (ALM, DES). 
High performance computing resources were provided by San Diego Supercomputing Cluster.

\bibliographystyle{mrp}
\bibliography{biblio}

\newpage
\appendix

\section{Derivation of $\alpha_\text{opt}$ in Method F} \label{app:alpha}

In this section, we derive the optimal coefficient $\alpha_\text{opt}$ in equation \eqref{eq:alpha_opt} that minimizes the variance of the inflated noise $\widetilde\eta$ in equation \eqref{eq:methodF_general}. First, notice that, since $\eta$ is independent of $\Q_\HF(\bm{x}) - \alpha Q^\dagger(\bm{x})$, we have
\begin{equation}
    \alpha_\text{opt} = \arg\min_{\alpha} \mathbb V^\mu[\widetilde\eta] =  \arg\min_{\alpha} \mathbb V^\mu[\Q_\HF(\bm{X}) - \alpha Q^\dagger(\bm{X})].
\end{equation}
Therefore, we need to find the value of $\alpha$ that minimizes the function 
\begin{equation}
\begin{aligned}
    \mathcal V(\alpha) &= \mathbb V^\mu[\Q_\HF(\bm{X}) - \alpha Q^\dagger(\bm{X})] \\
    &= \mathbb V^\mu[\Q_\HF(\bm{X})] -2\alpha \mathbb C^\mu[\Q_\HF(\bm{X}), Q^\dagger(\bm{X})] + \alpha^2 \mathbb V^\mu[Q^\dagger(\bm{X})],
\end{aligned}
\end{equation}
which is convex. Computing the zero of the first derivative, i.e. solving $\mathcal V'(\alpha_\text{opt}) = 0$, we obtain
\begin{equation}
    - 2 \mathbb C^\mu[\Q_\HF(\bm{X}), Q^\dagger(\bm{X})] + 2 \alpha_\text{opt} \mathbb V^\mu[Q^\dagger(\bm{X})] = 0,
\end{equation}
which implies
\begin{equation}
    \alpha_{\text{opt}} = \frac{\mathbb{C}^{\mu}\left[\Q_{\HF}(\bm{X}), Q^\dagger(\bm{X})  \right]}{\mathbb{V}^{\mu}\left[Q^\dagger(\bm{X}) \right]},
\end{equation}
which corresponds to equation \eqref{eq:alpha_opt}. \chloerev{Substituting $\alpha_{\text{opt}}$ into $\mathcal{V}(\alpha)$, we obtain:}

\begin{equation}  \label{eq:variance_opt}
    \chloerev{\mathcal{V}(\alpha_{\text{opt}}) = \mathbb{V}^{\mu}[\mathcal{Q}_{\text{HF}}(\bm{X})] (1 - \kappa^2),}
\end{equation}

\chloerev{where $\kappa$ is the Pearson correlation coefficient between $\mathcal{Q}_{\text{HF}}$ and $\mathcal{Q}^{\dagger}$:}

\begin{equation}
    \chloerev{\kappa = \frac{\mathbb{C}^{\mu}\left[\Q_{\HF}(\bm{X}), Q^\dagger(\bm{X})  \right]}{\sqrt{\mathbb{V}(\mathcal{Q}_{\text{HF}}(\bm{X})) \mathbb{V}(\mathcal{Q}^\dagger(\bm{X}))} }.}
\end{equation}

\chloerev{Equation \eqref{eq:variance_opt} quantifies the increase in variance relative to the observation noise $\eta$, and represents the minimum variance inflation achievable when replacing the high-fidelity model with a scaled version of its surrogate. Moreover, this variance is reduced as the correlation between $\mathcal Q_\HF$ and $Q^\dagger$ increases.}

\chloerev{This idea is inspired by generalized approximate control variate (ACV) estimators \cite{Gorodetsky_2020} and multifidelity Monte Carlo estimators \cite{peherstorfer2016optimal}. However, unlike in \cite{Gorodetsky_2020,peherstorfer2016optimal}, where the primary goal is to minimize the variance of the Monte Carlo estimator, our focus is on minimizing the variance of the inflated noise term, $\widetilde\eta$. The key point is that the high-fidelity model $\mathcal Q_\HF$ may, in some cases, be more strongly correlated with a scaled version of its surrogate $Q^\dagger$ than with the surrogate itself. In our approach, while replacing the high-fidelity model reduces computational cost, it also increases the variance of the posterior distribution due to the larger variance of the inflated noise. Therefore, introducing a scaling parameter $\alpha$, as done in \cite{Gorodetsky_2020,peherstorfer2016optimal}, provides a technique to mitigate this effect and potentially prevent the posterior variance from becoming too large. We finally remark that if $Q^\dagger$ is proportional to $\mathcal Q_\HF$, then the optimal choice is $\alpha_{\mathrm{opt}} = 1$, indicating that the scaling parameter $\alpha$ has no effect in this case.}



\newpage

\section{Two-dimensional analytical function with a larger dataset} \label{app:analytical_N_500}

\begin{figure}[!htb]
    \centering
    \begin{subfigure}[t]{0.75\textwidth}
        \centering
        \includegraphics[width=0.8\linewidth]{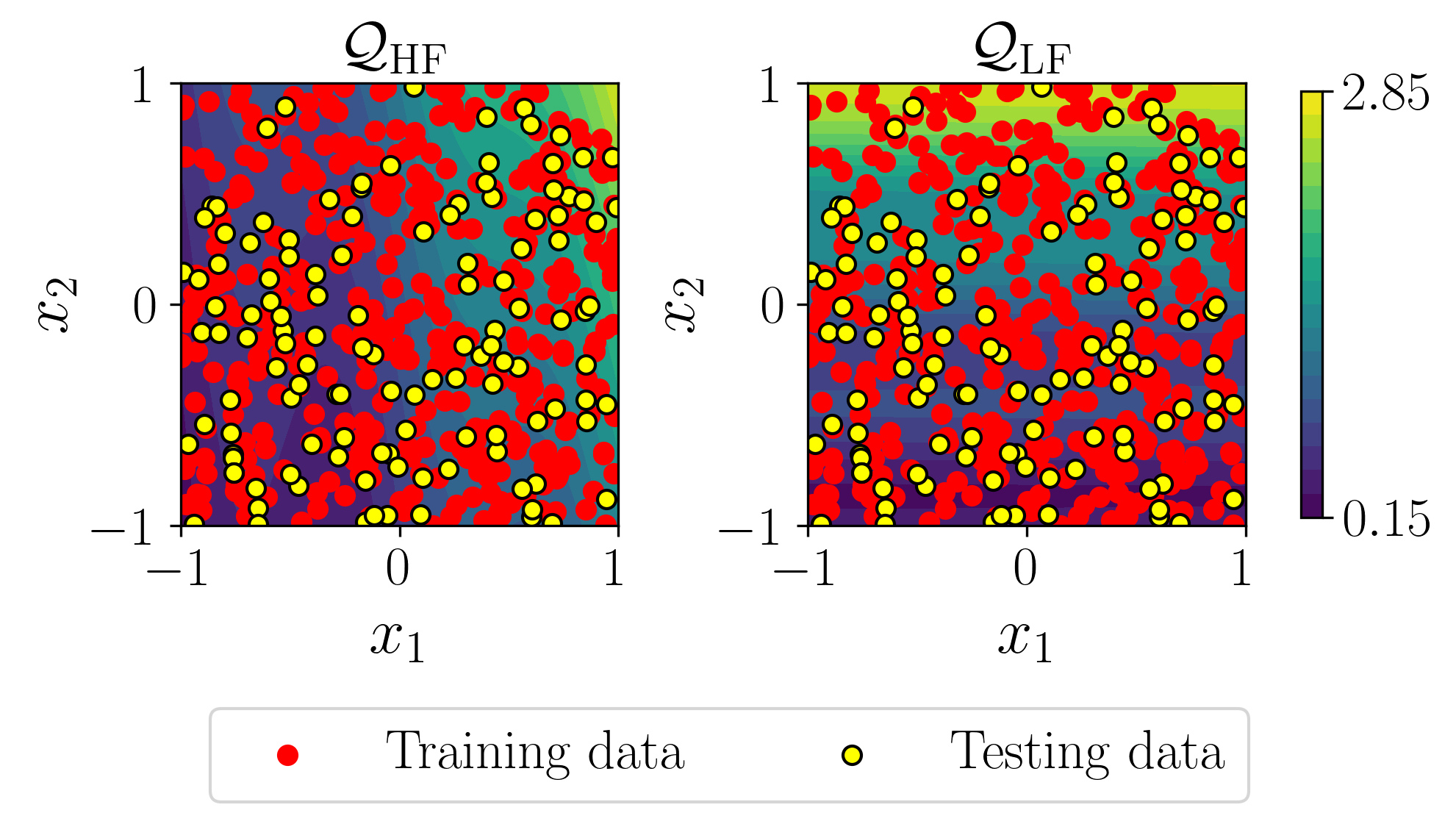}
        \caption{Training and testing data overlaid on the high- and low-fidelity models. \chloerev{Contour limits shown on right.}}
    \end{subfigure} 
    
    \vspace{1em}
    
    \begin{subfigure}[t]{0.85\textwidth}
        \centering
        \includegraphics[width=\textwidth]{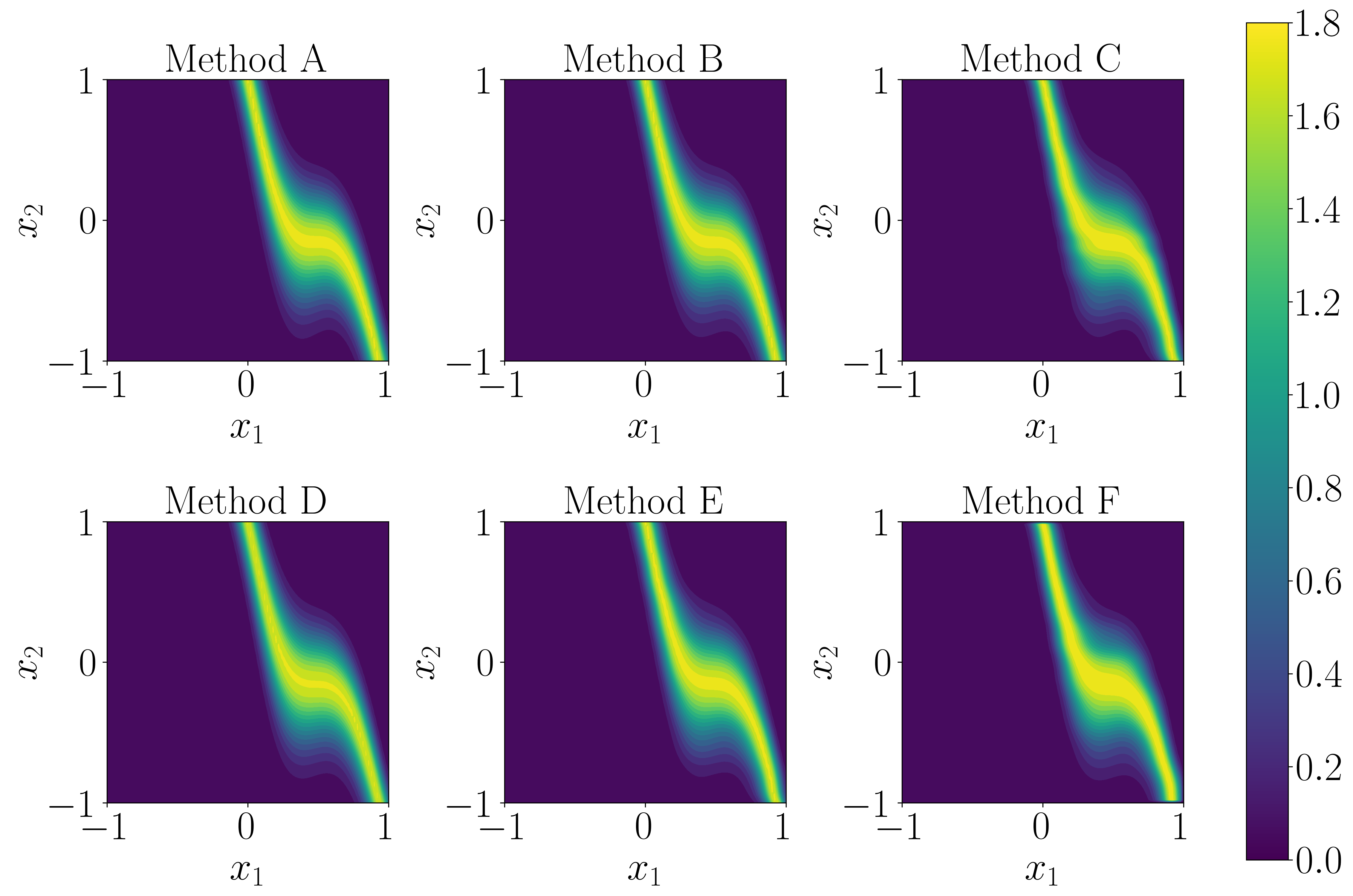}
        \caption{Posterior distributions computed using Methods A to F.}
    \end{subfigure}
    \caption{Posterior distributions for the two-dimensional analytical example. The surrogate models for Methods A to F are built using $N=500$ data points.}
    \label{fig:N500_app}
\end{figure}

We consider the model in \cref{sec:analytical_example} with the same setup, except for the number of data, which we now assume to be $N=500$, and report the numerical results in \cref{fig:N500_app}. We observe that when the number of data points is sufficiently large, all methods are able to accurately approximate the true posterior distribution.

\clearpage

\section{\chloerev{Benchmark function studies with Latin hypercube sampling}}\label{app:analytical_LHS}

\chloerev{We acknowledge that alternative space-filling designs of experiments have been applied to enhance the performance of multi-fidelity approaches, as in \cite{lee2024novel, chen2020optimization}. We previously used Latin hypercube sampling (LHS) and Sobol sequences for electrophysiology benchmarks \cite{SFA21, ZGS24b} and coronary circulation models \cite{seo2020multifidelity}. We consider Latin hypercube sampling (LHS) with the `maximin' criterion to generate 75 samples for training and 500 samples for testing. In this section, we confirm that LHS has little effect on the final results.}

\subsection{\chloerev{Two-dimensional analytical example}}

\begin{figure}[!htb]
    \centering
    \begin{subfigure}[t]{0.75\textwidth}
        \centering
        \includegraphics[width=0.8\linewidth]{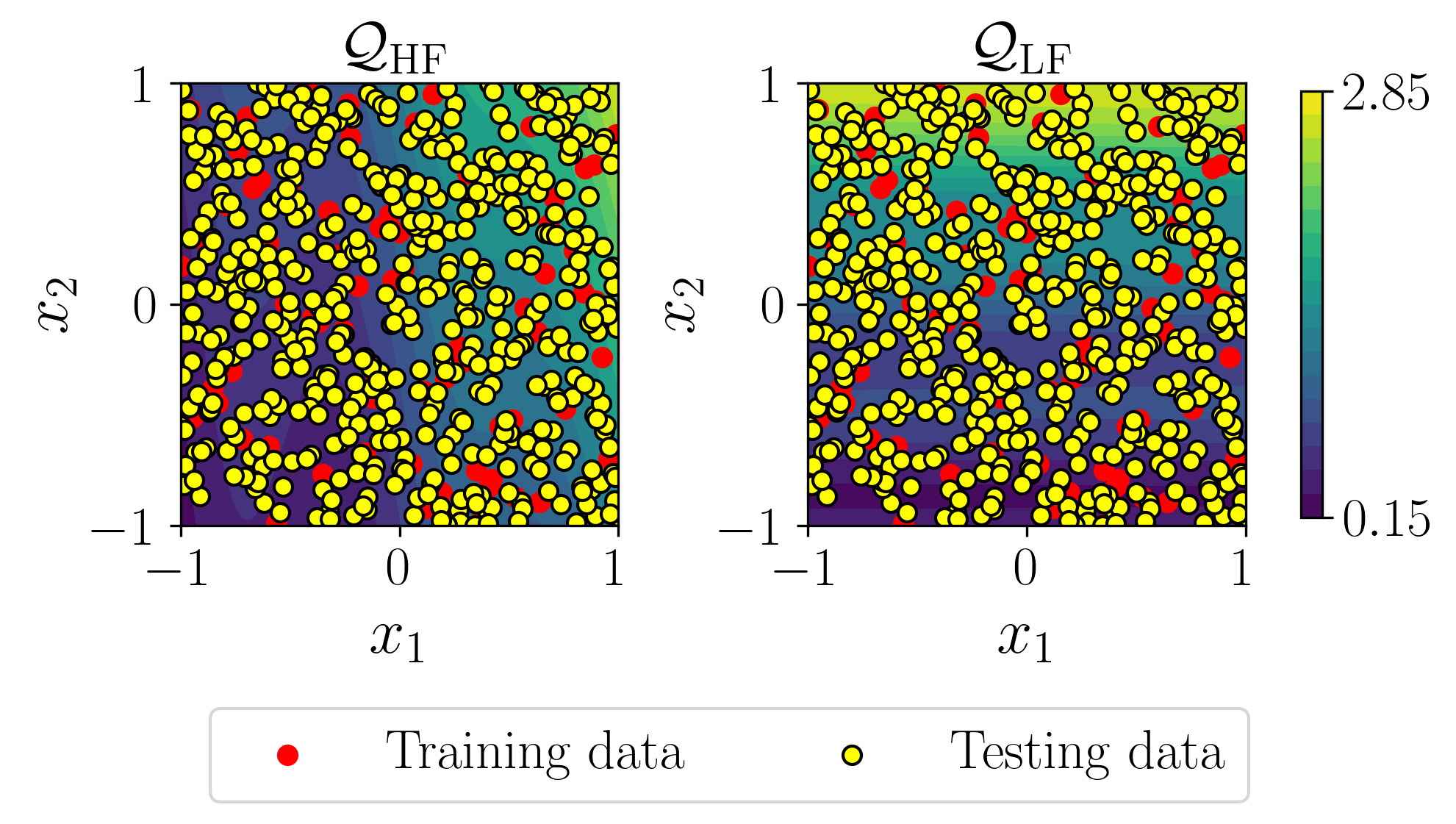}
        \caption{\chloerev{Training and testing data overlaid on the high- and low-fidelity models.} \chloerev{Contour limits shown on right.}}
    \end{subfigure} 
    
    \vspace{1em}
    
    \begin{subfigure}[t]{0.85\textwidth}
        \centering
        \includegraphics[width=\textwidth]{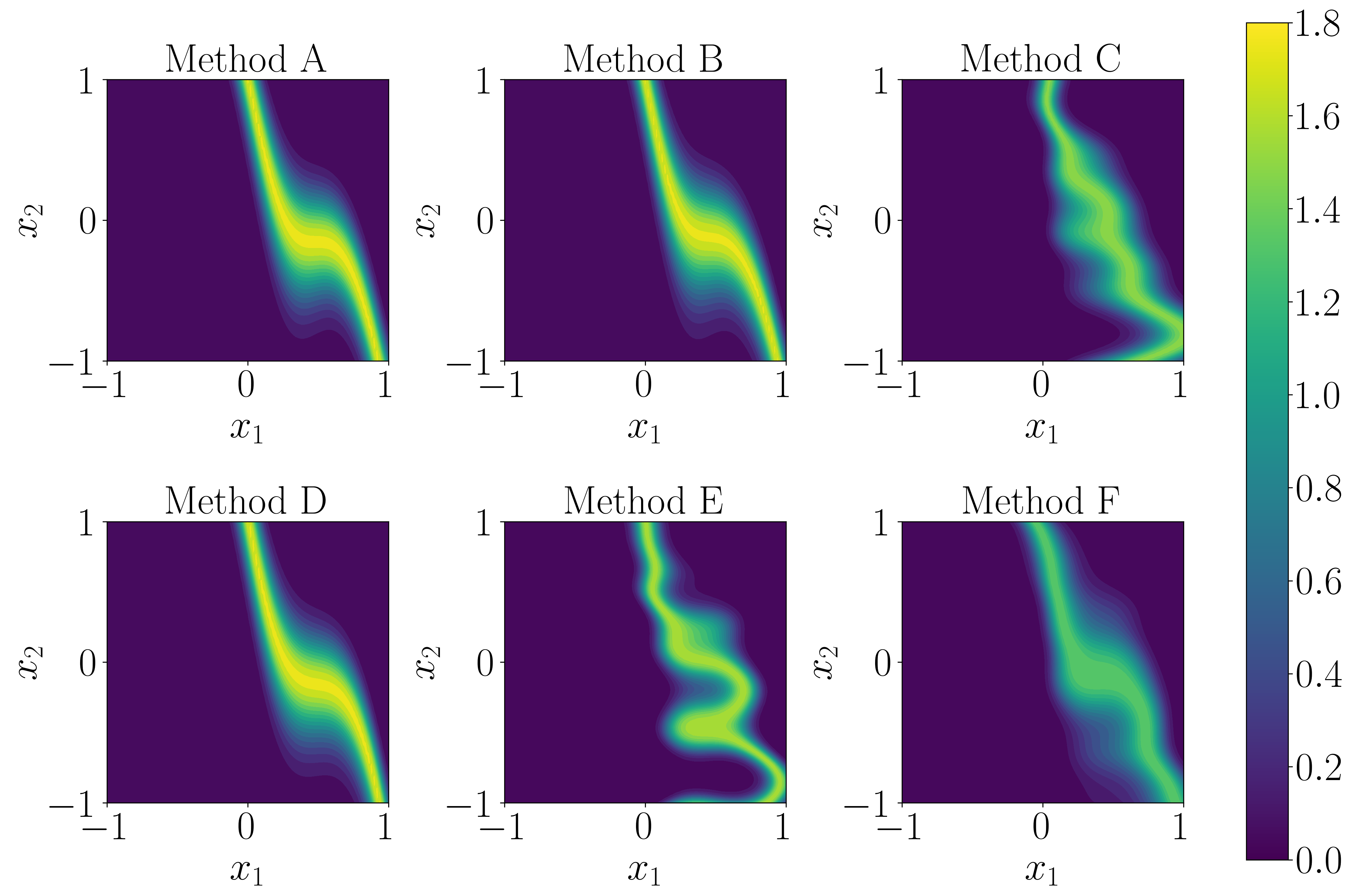}
        \caption{\chloerev{Posterior distributions computed using Methods A to F.}}
    \end{subfigure}
    
    \caption{\chloerev{Posterior distributions for the two-dimensional analytical example. The surrogate models for Methods A to F are built using $N=575$ data points.}}
    \label{fig:analytical_n_575}
\end{figure}

\chloerev{We display the training/testing data in Panel (a) and the resulting posterior distributions in Panel (b) of \cref{fig:analytical_n_575}. We quantify the Hellinger distance, trace, and cost of training the model and of evaluating the posterior in \cref{tab:analytical_N_575}. As before, we note that Method F has the largest variance and shows improvement from Method E.}

\begin{table}[!ht]
\centering
\caption{\chloerev{Accuracy, variance, and computational cost of the posterior distributions obtained using $N=575$ samples, for the two-dimensional analytical example. Results are averaged over 10 independent experiments. ``Posterior'' refers to the time spent to evaluate the posterior distribution on a $100 \times 100$ grid, one point at a time.}}
\resizebox{\textwidth}{!}{
\begin{tabular}{c c c c c}
\toprule
\multirow{2}{*}{\bf Method} & \multirow{2}{*}{\bf Hellinger distance} & \multirow{2}{*}{tr($\mathbb{C}^{\mu}(\cdot)$)} & \multicolumn{2}{c}{\bf Cost}\\
  & & & {\bf Optuna} & {\bf Posterior}\\
\midrule
 {\bf A} &  0.0 & $0.290$ & - & $0.55$ s\\
 {\bf B} &  $0.028 \pm 0.005$ & $\boldsymbol{0.288 \pm 0.002}$ & $1070 \pm 136$ s & $ 1.22 \pm 0.2$ s\\
 {\bf C} &  $0.238 \pm 0.077$ & $0.313 \pm 0.012$ & $\boldsymbol{891 \pm 92}$ s &  $1.0 \pm 0.2$ s\\
 {\bf D} &  $\boldsymbol{0.027 \pm 0.007}$ & $0.291 \pm 0.001$ & $3500 \pm 338$ s & $\boldsymbol{1.0 \pm 0.1}$ s\\
 {\bf E} &  $0.254 \pm 0.040$ & $0.304 \pm 0.020$ & $3472 \pm 406$ s & $1.1 \pm 0.1$ s\\
 {\bf F} & $0.252 \pm 0.044$ & $0.332 \pm 0.012$ & $6281 \pm 886$ s & $9.8 \pm 6.2$ s\\
 \bottomrule
\end{tabular}
}
\label{tab:analytical_N_575}
\end{table}

\clearpage

\subsection{\chloerev{Michalewicz example}}

\begin{figure}[!htb]
    \centering
    \begin{subfigure}[t]{0.75\textwidth}
        \centering
        \includegraphics[width=0.8\linewidth]{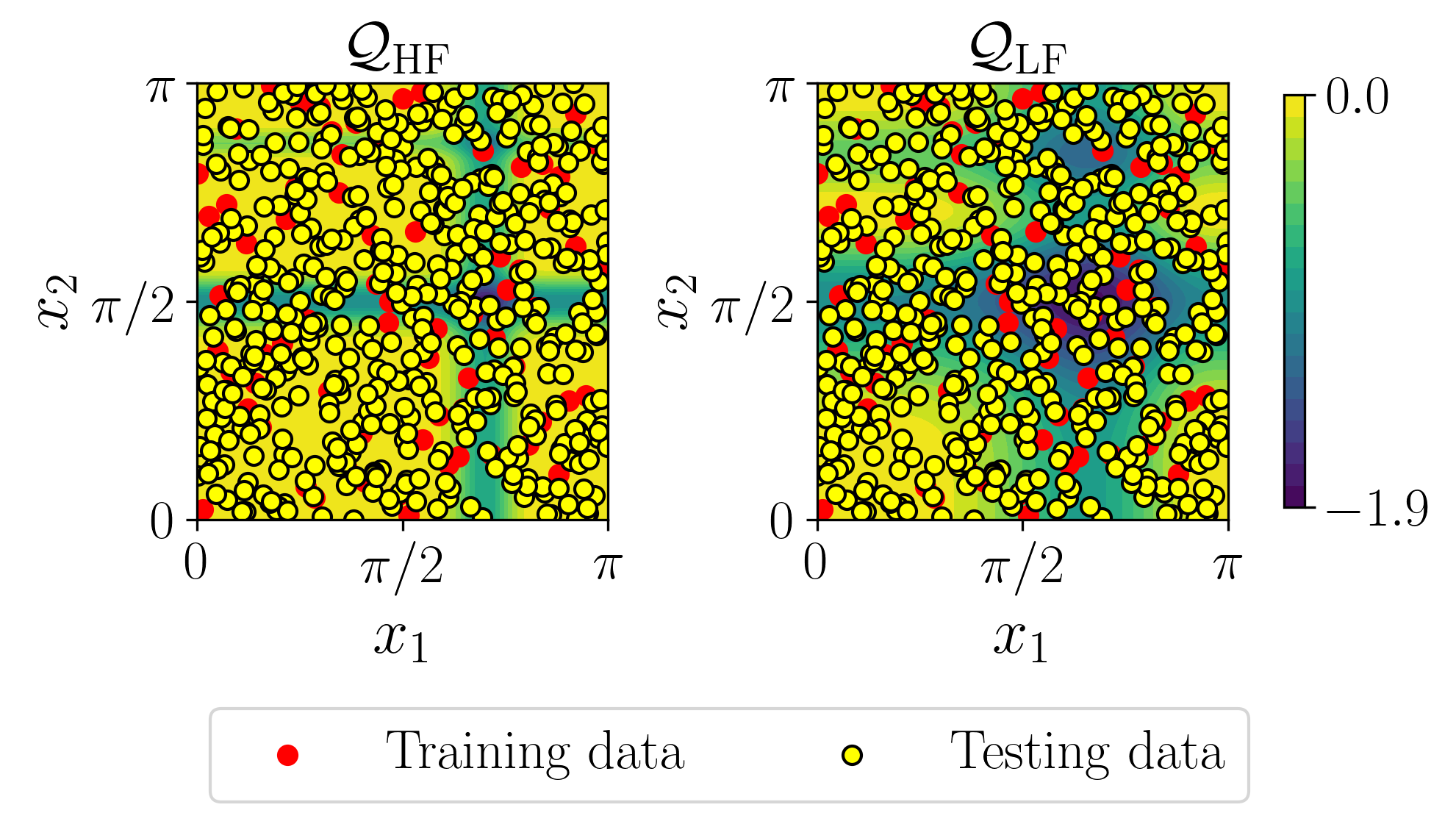}
        \caption{\chloerev{Training and testing data overlaid on the high- and low-fidelity models.} \chloerev{Contour limits shown on right.}}
    \end{subfigure} 
    
    \vspace{1em}
    
    \begin{subfigure}[t]{0.85\textwidth}
        \centering
        \includegraphics[width=\textwidth]{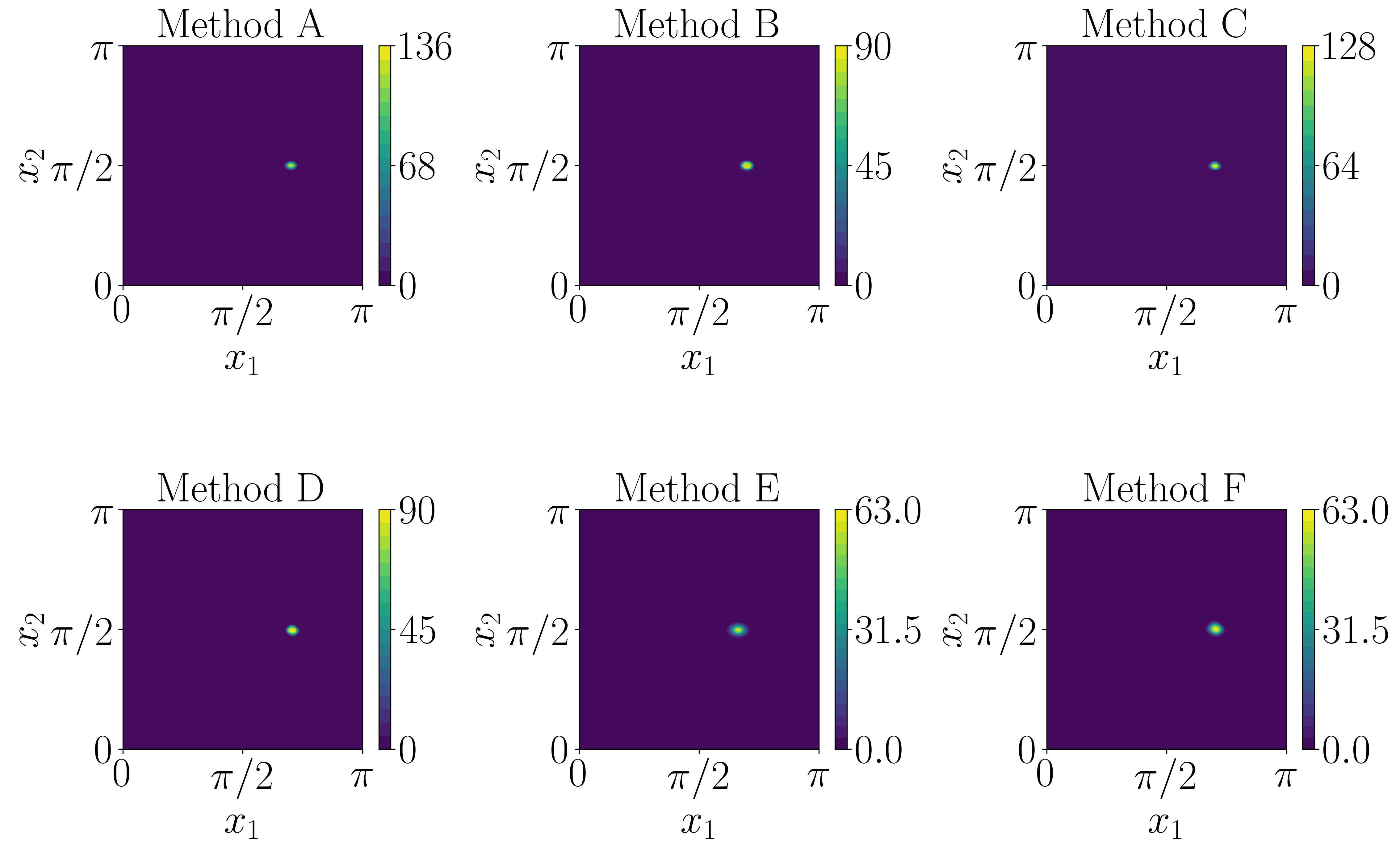}
        \caption{\chloerev{Posterior distributions obtained using the six different methods.}}
    \end{subfigure}
    
    \caption{\chloerev{Posterior distributions for the Michalewicz function test case, where the surrogate models are built using $N=575$ data.}}
    \label{fig:michalewicz_n_575}
\end{figure}

\chloerev{We display the training/testing data in Panel (a) and the resulting posterior distributions in Panel (b) of \cref{fig:michalewicz_n_575}. We quantify the Hellinger distance, trace, and cost of training the model and of evaluating the posterior in \cref{tab:grad_N_575}. As before, we note that there is a significant improvement in the Hellinger distance from Method E to Method F at the cost of increased variance, indicating that this methodology works best for sharp discrepancies.}

\begin{table}[!ht]
\centering
\caption{\chloerev{Accuracy, variance, and computational cost of the posterior distributions obtained using $N=575$ samples, for the Michalewicz function test case. Results are averaged over 10 independent experiments.}}
\resizebox{\textwidth}{!}{
\begin{tabular}{c c c c c}
\toprule
\multirow{2}{*}{\bf Method} & \multirow{2}{*}{\bf Hellinger distance} & \multirow{2}{*}{tr($\mathbb{C}^{\mu}(\cdot)$)} & \multicolumn{2}{c}{\bf Cost}\\
& & & {\bf Optuna} & {\bf Posterior}\\
\midrule
{\bf A} &  0.0 & 0.001786 & - & 0.55 s\\
{\bf B} &  $\bm{0.413 \pm 0.211}$ & $0.001624 \pm 0.001037$ & $813 \pm 94$ s & $\boldsymbol{0.7 \pm 0.1}$ s\\
{\bf C} &  $0.450 \pm 0.212$ & $\bm{0.001114 \pm 0.000630}$ & $\boldsymbol{783 \pm 81}$ s & $1.1 \pm 0.2$ s\\
{\bf D} &  $0.492 \pm 0.200$ & $0.001323 \pm 0.000589$ & $4197 \pm 329$ s & $0.8 \pm 0.1$ s\\
{\bf E} &  $0.869 \pm 0.087$ & $0.01615 \pm 0.01263$ & $4109 \pm 466$ s & $1.2 \pm 0.2$ s\\
{\bf F} & $0.583 \pm 0.1393$ & $0.01784 \pm 0.01573$ & $6810 \pm 981$ s & $11.8 \pm 10.0$  s\\
\bottomrule
\end{tabular}}
\label{tab:grad_N_575}
\end{table}

\clearpage

\subsection{\chloerev{Corresponding hyperparameters}}

\chloerev{The corresponding hyperparameters for these additional experiments are:}

\begin{table}[!ht]
    \centering
    \caption{\chloerev{Hyperparameters selected via Optuna for the normalizing flow part of Method F, in the \chloerev{five} numerical experiments. The parameters are: number of layers in the normalizing flow $\ell_\NF$, number of neurons per layer in the normalizing flow $n_\NF$, learning rate $h$ and exponential scheduler step $\zeta$.}}
    \begin{tabular}{c|ccccc}
    \toprule
         Example & $\ell_\NF$ & $n_\NF$ & $b$ & $h$ & $\zeta$ \\
    \midrule
        Analytical & 10 & 12 & 2 & 6.10e-5 & 0.99910 \\
        Michalewicz & 6 & 7 & 2 & 7.92e-5 & 0.99945 \\
    \bottomrule    
    \end{tabular}
    \label{tab:LHS_hyperparameters_NF}
\end{table}

\begin{table}[!ht]
    \centering
    \caption{\chloerev{Hyperparameters selected via Optuna for the surrogate modeling part of all the methods, in the \chloerev{five} numerical experiments. The parameters are: number of layers in the surrogate $\ell_\S$, number of neurons per layer in the surrogate $n_\S$, number of layers in the autoencoder $\ell_{\E/\D}$, number of neurons per layer in the autoencoder $n_{\E/\D}$, learning rate $h$ and exponential scheduler step $\zeta$.}}
    \begin{tabular}{c|cccccc}
        \multicolumn{7}{c}{\textbf{Analytical function}} \\
    \toprule
         {\bf Method} & $\ell_\S$ & $n_\S$ & $\ell_{\E/\D}$ & $n_{\E/\D}$ & $h$ & $\zeta$ \\
    \midrule
        {\bf A} & - & - & - & - & - & - \\
        {\bf B} & 10 & 17 & - & - & 0.0009826 & 0.99988 \\
        {\bf C} & 5 & 15 & - & - & 0.0009497 & 0.99970 \\
        {\bf D} & 1 & 10 & 3 & 18 & 0.0009211 & 0.99986 \\
        {\bf E} & 1 & 4 & 4 & 9 & 0.0007181 & 0.99924 \\
        {\bf F} & 1 & 5 & 3 & 19 & 0.00099474 & 0.99976 \\
    \bottomrule    
    \end{tabular}

    \vspace{0.5cm}

    \begin{tabular}{c|cccccc}
        \multicolumn{7}{c}{\textbf{Michalewicz function}} \\
    \toprule
         {\bf Method} & $\ell_\S$ & $n_\S$ & $\ell_{\E/\D}$ & $n_{\E/\D}$ & $h$ & $\zeta$ \\
    \midrule
        {\bf A} & - & - & - & - & - & - \\
        {\bf B} & 3 & 17 & - & - & 0.0007801 & 0.99959 \\
        {\bf C} & 3 & 13 & - & - & 0.0007137 & 0.99988 \\
        {\bf D} & 1 & 14 & 2 & 11 & 0.0006610 & 0.99958 \\
        {\bf E} & 1 & 12 & 2 & 12 & 0.0008506 & 0.99943 \\
        {\bf F} & 1 & 16 & 3 & 20 & 0.0008948 & 0.99934 \\
    \bottomrule    
    \end{tabular}
    \label{tab:LHS_hyperparameters_all}
\end{table}

\clearpage

\section{Hyperparameter tuning} \label{app:optuna_hyperparams}

The hyperparameters of the neural networks used in the numerical examples are tuned using the Optuna optimization framework~\cite{Optuna}, over 100 iterations on a single CPU. The dataset was split into 25\% for testing and 75\% for training. The hyperparameters included in the optimization procedure, along with their admissible ranges, are listed in \cref{tab:appendix_optuna_hyperparams}. In particular, we consider number of layers $\ell$, number of neurons per layer $n$, learning rate $h$, exponential scheduler step $\zeta$, and number of blocks $b$ in the normalizing flow. All neural networks have $\mathrm{tanh}$ activation function and are trained for 10,000 epochs using the Adam optimizer \cite{adam_opt_2017} with a weight decay of 2e-4. The hyperparameters selected via Optuna for the normalizing flow and surrogate modeling steps, which are used in the numerical experiments, are reported in \cref{tab:hyperparameters_NF,tab:hyperparameters_all_analytic,tab:hyperparameters_all_cardio}, respectively.

\begin{table}[!ht]
    \centering
    \caption{Hyperparameters and corresponding admissible ranges tuned using Optuna}
    \begin{tabular}{c c}
        \toprule
        {\bf Hyperparameter} & {\bf Range}\\
        \midrule
        Layers $\ell$ & \{1, \dots, 10\} \\
        Neurons per layer $n$ & \{1, \dots, 20\} \\
        Learning rate $h$ & [1e-3, 1e-5] \\
        Scheduler step $\zeta$ & [0.999, 0.9999] \\
        Blocks $b$ (NF) & \{1, \dots, 4\} \\
        \bottomrule
    \end{tabular}
    \label{tab:appendix_optuna_hyperparams}
\end{table}

\begin{table}[!ht]
    \centering
    \caption{Hyperparameters selected via Optuna for the normalizing flow part of Method F, in the \chloerev{five} numerical experiments. The parameters are: number of layers in the normalizing flow $\ell_\NF$, number of neurons per layer in the normalizing flow $n_\NF$, learning rate $h$ and exponential scheduler step $\zeta$.}
    \begin{tabular}{c|ccccc}
    \toprule
         Example & $\ell_\NF$ & $n_\NF$ & $b$ & $h$ & $\zeta$ \\
    \midrule
        Analytical & 9 & 2 & 2 & 0.0004489 & 0.99970 \\
        Michalewicz & 1 & 1 & 3 & 0.0009722 & 0.99916 \\
        \chloerev{Borehole} & \chloerev{7} & \chloerev{1} & \chloerev{2} & \chloerev{0.0002005} & \chloerev{0.99908} \ \\
        Circuit & 4 & 1 & 3 & 0.0007937 & 0.99906\ \\
        Aorto-iliac & 6 & 2 & 2 & 0.0009964 & 0.99909 \\
    \bottomrule    
    \end{tabular}
    \label{tab:hyperparameters_NF}
\end{table}

\begin{table}[!ht]
    \centering
    \caption{Hyperparameters selected via Optuna for the surrogate modeling part of all the methods, in the \chloerev{three analytical test cases}. The parameters are: number of layers in the surrogate $\ell_\S$, number of neurons per layer in the surrogate $n_\S$, number of layers in the autoencoder $\ell_{\E/\D}$, number of neurons per layer in the autoencoder $n_{\E/\D}$, learning rate $h$ and exponential scheduler step $\zeta$.}
    
    \begin{tabular}{c|cccccc}
        \multicolumn{7}{c}{\textbf{Analytical function}} \\
    \toprule
         {\bf Method} & $\ell_\S$ & $n_\S$ & $\ell_{\E/\D}$ & $n_{\E/\D}$ & $h$ & $\zeta$ \\
    \midrule
        {\bf A} & - & - & - & - & - & - \\
        {\bf B} & 4 & 20 & - & - & 0.000916 & 0.99975 \\
        {\bf C} & 6 & 20 & - & - & 0.000736 & 0.99956 \\
        {\bf D} & 1 & 5 & 3 & 15 & 0.000954 & 0.9998 \\
        {\bf E} & 6 & 20 & 3 & 20 & 0.000576 & 0.99984 \\
        {\bf F} & 1 & 4 & 3 & 10 & 0.000835 & 0.99982 \\
    \bottomrule    
    \end{tabular}

    \vspace{0.5cm}

    \begin{tabular}{c|cccccc}
        \multicolumn{7}{c}{\textbf{Michalewicz function}} \\
    \toprule
         {\bf Method} & $\ell_\S$ & $n_\S$ & $\ell_{\E/\D}$ & $n_{\E/\D}$ & $h$ & $\zeta$ \\
    \midrule
        {\bf A} & - & - & - & - & - & - \\
        {\bf B} & 5 & 19 & - & - & 0.0007474 & 0.99979 \\
        {\bf C} & 1 & 16 & - & - & 0.0003312 & 0.99953 \\
        {\bf D} & 3 & 7 & 2 & 9 & 0.0009680 & 0.99958 \\
        {\bf E} & 1 & 3 & 7 & 9 & 0.0004689 & 0.99952 \\
        {\bf F} & 1 & 16 & 3 & 6 & 0.0005113 & 0.99915 \\
    \bottomrule    
    \end{tabular}

    \vspace{0.5cm}

    \begin{tabular}{c|cccccc}
        \multicolumn{7}{c}{\textbf{Borehole example}} \\
    \toprule
         {\bf Method} & $\ell_\S$ & $n_\S$ & $\ell_{\E/\D}$ & $n_{\E/\D}$ & $h$ & $\zeta$ \\
    \midrule
        {\bf A} & - & - & - & - & - & - \\
        {\bf B} & \chloerev{4} & \chloerev{16} & - & - & \chloerev{0.0009997} & \chloerev{0.99984} \\
        {\bf C} & \chloerev{3} & \chloerev{18} & - & - & \chloerev{0.0009421} & \chloerev{0.99986} \\
        {\bf D} & \chloerev{3} & \chloerev{20} & \chloerev{9} & \chloerev{17} & \chloerev{0.0009005}  & \chloerev{0.99900} \\
        {\bf E} & \chloerev{2} & \chloerev{4} & \chloerev{5} & \chloerev{15} & \chloerev{0.0008717} & \chloerev{0.99987} \\
        {\bf F} & \chloerev{2} & \chloerev{4} & \chloerev{5} & \chloerev{15} & \chloerev{0.0008717} & \chloerev{0.99987}  \\
    \bottomrule    
    \end{tabular}

    \label{tab:hyperparameters_all_analytic}
\end{table}

\begin{table}[!ht]
    \centering
    \caption{Hyperparameters selected via Optuna for the surrogate modeling part of all the methods, \chloerev{in the two cardiovascular examples}. The parameters are: number of layers in the surrogate $\ell_\S$, number of neurons per layer in the surrogate $n_\S$, number of layers in the autoencoder $\ell_{\E/\D}$, number of neurons per layer in the autoencoder $n_{\E/\D}$, learning rate $h$ and exponential scheduler step $\zeta$.}

    \begin{tabular}{c|cccccc}
        \multicolumn{7}{c}{\textbf{Circuit example}} \\
    \toprule
         {\bf Method} & $\ell_\S$ & $n_\S$ & $\ell_{\E/\D}$ & $n_{\E/\D}$ & $h$ & $\zeta$ \\
    \midrule
        {\bf A} & - & - & - & - & - & - \\
        {\bf B} & 1 & 17 & - & - & 0.0007986 & 0.99985 \\
        {\bf C} & 5 & 19 & - & - & 0.0008066 & 0.99981 \\
        {\bf D} & 1 & 8 & 4 & 16 & 0.0009245 & 0.99987 \\
        {\bf E} & 2 & 18 & 2 & 19 & 0.0006313 & 0.99979 \\
        {\bf F} & 4 & 17 & 3 & 20 & 0.0006218 & 0.99985 \\
    \bottomrule    
    \end{tabular}

    \vspace{0.5cm}

    \begin{tabular}{c|cccccc}
        \multicolumn{7}{c}{\textbf{Aorto-iliac example}} \\
    \toprule
         {\bf Method} & $\ell_\S$ & $n_\S$ & $\ell_{\E/\D}$ & $n_{\E/\D}$ & $h$ & $\zeta$ \\
    \midrule
        {\bf A} & - & - & - & - & - & - \\
        {\bf B} & 7 & 15 & - & - & 0.0003957 & 0.99986 \\
        {\bf C} & 4 & 16 & - & - & 7.99e-5 & 0.99911 \\
        {\bf D} & 2 & 2 & 4 & 19 & 0.000952 & 0.99973 \\
        {\bf E} & 3 & 2 & 4 & 20 & 1.01e-4 & 0.99926 \\
        {\bf F} & 2 & 4 & 1 & 19 & 0.000166 & 0.99914 \\
    \bottomrule    
    \end{tabular}

    \label{tab:hyperparameters_all_cardio}
\end{table}

\end{document}